%% file: paper.tex
\documentclass[]{jingdong}
\usepackage{colortbl}
\usepackage[toc,page,header]{appendix}

\input{math_commands.tex}

\usepackage{amsthm}
\usepackage{float}
\usepackage{fontawesome}
\newcommand{\hficon}{\raisebox{-0.18\height}{\includegraphics[height=1.02em]{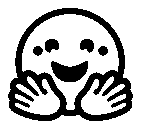}}}

\AtBeginDocument{%
    \crefname{figure}{Fig.}{Figs.}%
    \Crefname{figure}{Fig.}{Figs.}%
    \crefname{table}{Tab.}{Tabs.}%
    \Crefname{table}{Tab.}{Tabs.}%
}

\theoremstyle{plain}

\theoremstyle{definition}

\theoremstyle{remark}

\newtcolorbox{takeaway}{
    colback=black!3, colframe=black!55,
    boxrule=0.6pt, arc=1pt, boxsep=0pt,
    left=6pt, right=6pt, top=4pt, bottom=4pt,
    before skip=6pt, after skip=6pt
}

\AddToHook{cmd/@makecol/before}{%
    \ifnum\value{page}=\getpagerefnumber{fig:intro}\relax
        \raggedbottom
    \fi
}

\title{Harness Tax for GUI Agents}

\author[\ddagger]{Ji~Zhao}
\author[\ddagger]{Taiqiang~Wu} 
\author{Xinlei~Yu}
\author{Caoyuan~Ma}
\author{Haiyuan~Wan}
\author{Wenzheng~Chang}
\author{Hongyang~Li}
\author{Shuai~Dong}
\author{Wenpu~Liu}
\author{Jinghao~Zhang}
\author{Zheming~Liang}
\author{Yuqi~Xu}
\author{Ziyue~Wang}
\author{Haowen~Hou}
\author{Dianyi~Wang}
\author{Jun~Yu}
\author{Yufei~Gu}
\author{Zeke~Xie}
\author{Ngai~Wong}
\author{Xuehui~Wang}
\author{Yongfu~Zhu}
\author[\dagger]{Jiaqi~Wang}
\author[*]{Wenqi~Shao}

\affiliation{USTC \quad HKU \quad CUHK \quad UTokyo \quad THU \quad SJTU \quad FDU \quad PKU \quad HKUST-GZ \quad SII \quad Explore Academy, JD}

\contribution[\ddagger]{Equal contribution}
\contribution[\dagger]{Project leader}
\contribution[*]{Corresponding author}

\headerlink{https://masslab-sii.github.io/Harness2/}{\faGlobe}{Project}
\headerlink{https://github.com/MassLab-SII/Harness2}{\faGithub}{Code}
\headerlink{https://huggingface.co/papers/2611.36377}{\hficon}{Paper}

\abstract{
Graphical user interface (GUI) agents automate tasks across desktop applications through a harness that mediates observations, actions, and interaction history.
Despite growing emphasis on increasingly challenging GUI tasks and stable performance across heterogeneous harnesses, cross-harness robustness has yet to be systematically benchmarked.
In this paper, we first reveal a \textbf{Harness Tax}: substantial performance degradation when a fixed model is switched from a customized harness to another.
To investigate this sensitivity, we introduce \textit{$Harness^2$}, a configurable and composable evaluation framework supporting 126 harness configurations across action interfaces, output protocols, and observation--history representations.
We also propose the \textbf{Harness Index}~(HI) to jointly summarize mean performance and cross-harness variability.
Experiments with eight open-weight and closed-source models reveal model-specific interface preferences and lower mean scores under history compression.
In an exploratory comparison on 55 fixed tasks, changing only the action interface produces score shifts exceeding 50 percentage points in opposite directions for two models, reversing their ranking.
We further examine two mitigation strategies: validation feedback allows the model to recover from the vast majority of format errors within the retry limit, and supervised fine-tuning improves mean scores across all evaluated component settings.
We hope these findings motivate evaluation and training across diverse harnesses as a step toward harness-agnostic GUI agents and reliable deployment.
}

\begin{document}
\maketitle

\begin{figure}[!t]
    \centering
    \includegraphics[width=\linewidth]{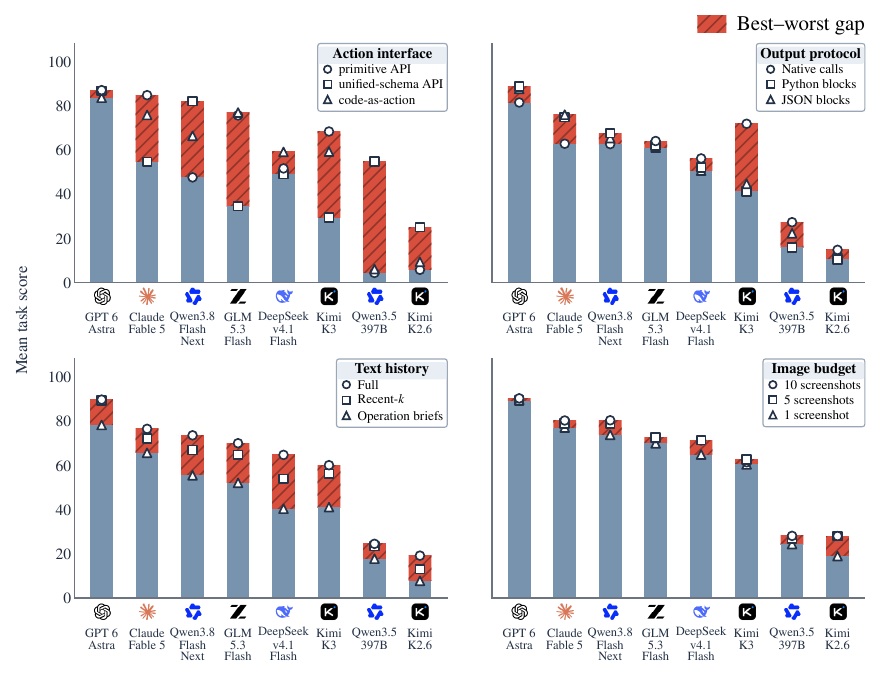}
    \caption{\textbf{Harness choices produce model-specific score gaps.}
    For each component, we average each setting's mean original task score over matched groups using the controls in Sec.~\ref{sec:component-sensitivity}.
    The blue bar shows the lowest setting mean, and the red hatched extension shows the gap to the highest setting mean.
    Open marker shapes identify the settings. 
    Models are ordered by mean task score.}
    \label{fig:intro}
\end{figure}

\section{Introduction}
\label{sec:intro}

Graphical user interface (GUI) agents translate natural-language requests into computer workflows, enabling users to delegate tasks across desktop and web applications~\citep{agashe2025agents,qin2025uitars,xu2024aguvis,xue2026evocua}.
For these capabilities to be reusable across systems, models must operate effectively through different interaction interfaces.
Such interaction is mediated by an \emph{agent harness}: the software layer that presents observations, exposes executable actions, and organizes interaction history~\citep{xie2024osworld,chezelles2025browsergym,agashe2026evolving}.
The harness therefore shapes both the evidence available to a model and the way its decisions are translated into executable actions.

This dependence raises a central question for general-purpose computer use: \emph{how reliably can a model solve the same tasks under different harnesses?}
Harnesses may express the same computer operations through different action schemas and output protocols, and expose different amounts or representations of visual and interaction history.
These choices matter when integrating a model into systems with different tool APIs and context budgets ~\citep{chezelles2025browsergym}.
For deployment, understanding their impact helps determine whether an agent's capabilities can be reused in a new system.
For evaluation, it determines how broadly benchmark results can inform model selection.
A score obtained under a single harness characterizes a particular model--harness pairing; whether that performance, or the resulting model ranking, carries over to another harness remains an empirical question.
\emph{Cross-harness robustness} is thus an important dimension of GUI-agent capability that requires explicit evaluation.

Existing benchmarks have established a foundation for measuring GUI-agent performance.
OSWorld and Windows Agent Arena evaluate agents on desktop and web workflows~\citep{xie2024osworld,bonatti2024windowsagentarena}, while more recent benchmarks such as OSWorld2.0 and Agents' Last Exam (ALE) extend evaluation toward more complex and longer-horizon tasks~\citep{yuan2026osworld2,sun2026agentslastexam}.
These advances broaden task coverage and difficulty.
A complementary question is how performance on those tasks depends on the interaction setup.
Although model developers increasingly emphasize robustness and generalization across agent harnesses~\citep{qwen2026flashnext,deepseek2026v41flash,kimi2026k3}, systematic evaluation under controlled changes to action interfaces, output protocols, and interaction histories remains limited.
Assessing progress toward general-purpose GUI agents therefore requires examining both the tasks they can complete and the interaction conditions under which their performance holds.

In this paper, we first reveal a \textbf{Harness Tax}: substantial performance degradation when a fixed model is switched from a customized harness to another.
Fig.~\ref{fig:intro} summarizes this sensitivity: the best-worst score gap and the preferred option vary across models and harness components.
For example, changing the harness while keeping the model and tasks fixed reduces Kimi K3's mean score from 87.66\% to 32.72\%.
The same change raises Qwen3.8 Flash Next's mean score from 28.90\% to 81.64\%, reversing the models' ranking.
This reversal makes systematic cross-harness evaluation necessary for reliable model comparisons.

To systematically evaluate such sensitivity, we introduce \textit{$Harness^2$}, to the best of our knowledge, the first configurable, composable framework for cross-harness robustness evaluation of desktop GUI agents.
We vary \emph{interaction format} through three action interfaces and three output protocols, and \emph{observation--history representation} through history retention, screenshot count, and input resolution.
Supported choices include native tool calls, Python or JSON blocks, and histories ranging from full records to recent turns with briefs of earlier operations or operation briefs alone.
Valid combinations of these choices and screenshot budgets of 1, 5, or 10 images yield 126 harness configurations.
We also propose the \textbf{Harness Index}~(HI) for benchmarking, which combines mean performance and cross-harness variability.

We evaluate eight open-weight and closed-source models across all 126 harness configurations.
Within-model comparisons reveal pronounced interface dependence: models with high scores under favorable configurations can suffer large losses after an interface change.
Output-format sensitivity varies by model, while replacing full histories with operation briefs lowers mean scores across the evaluated models.
For Qwen3.8 Flash Next, this change reduces the mean score from 73.48\% to 55.41\% with only the current screenshot.

\noindent To summarize, our contributions are threefold:

\begin{itemize}
\item \textbf{Harness Tax.} We identify substantial performance degradation and model-ranking reversals when the same models are evaluated under different harnesses.
\item \textbf{Configurable and Composable Harness Evaluation.}
We introduce \textit{$Harness^2$}, to the best of our knowledge, the first configurable, composable framework for cross-harness evaluation of desktop GUI agents, with 126 harness configurations and HI to jointly assess performance and cross-harness stability.

\item \textbf{Empirical Analysis.} We characterize model-specific component sensitivity and failure mechanisms, and investigate mitigation through validation feedback and post-training.
\end{itemize}

\section{Related Work}
\label{sec:related}
\paragraph{GUI Agent Models.}
Native GUI agents learn perception and action policies from demonstrations and interaction experience~\citep{qin2025uitars,xu2024aguvis,wang2026opencua,xue2026evocua}.
SeeClick, OS-ATLAS, ShowUI, and ShowUI-$\pi$ address visual grounding and fine-grained interaction, which are essential to successful computer use~\citep{cheng2024seeclick,wu2024osatlas,lin2024showui,hu2026showuipi}.
Modular systems such as Agent S, Agent S2, and Agent S3 combine model capabilities with specialized interfaces and inference-time planning~\citep{agashe2025agents,agashe2025agents2,gonzalezpumariega2025agents3}.
Tool integration and self-improvement for general computer agents are investigated in OS-Copilot~\citep{wu2024oscopilot}, and recursive improvement of data and models in CUA-Master~\citep{yu2026cuamaster}.

\paragraph{Task and Tool-Use Benchmarks.}
OSWorld and Windows Agent Arena evaluate desktop workflows~\citep{xie2024osworld,bonatti2024windowsagentarena}, while WebArena, VisualWebArena, and AndroidWorld cover web and mobile environments~\citep{zhou2023webarena,koh2024visualwebarena,rawles2024androidworld}.
WorkArena and WorkArena++ target enterprise workflows and compositional planning tasks~\citep{drouin2024workarena,boisvert2024workarenaplus}.
AgentBench and ToolBench broaden evaluation to heterogeneous agent environments and real-world APIs~\citep{liu2023agentbench,qin2023toolllm}.
Whereas these benchmarks vary the tasks under a given interaction setup, we keep the desktop tasks fixed and vary the harness through which the model interacts with them.

\paragraph{Harness Design and Evaluation Reliability.}
ReAct and SWE-agent illustrate the importance of reasoning--action loops and agent--computer interface design~\citep{yao2023react,yang2024sweagent}; Agent S4 explicitly evolves interfaces from task experience~\citep{agashe2026evolving}.
BrowserGym and AgentLab provide configurable environments for web-agent research~\citep{chezelles2025browsergym}, and OSWorld also examines several observation and history settings~\citep{xie2024osworld}.
FormatSpread shows that even surface-level prompt choices can alter measured performance~\citep{sclar2023formatspread}.
StableToolBench addresses instability in tool evaluation, while $\tau$-bench and $\tau^2$-bench study reliability in stateful interactions between agents and tools~\citep{guo2024stabletoolbench,yao2024taubench,barres2025tau2bench}.
In contrast, we systematically combine the components of desktop harnesses and aggregate performance at the task level.

\section{Harness Tax: Degradation across Harnesses}
\label{sec:harness-tax}
\begin{figure}[t]
    \centering
    \includegraphics[width=\linewidth]{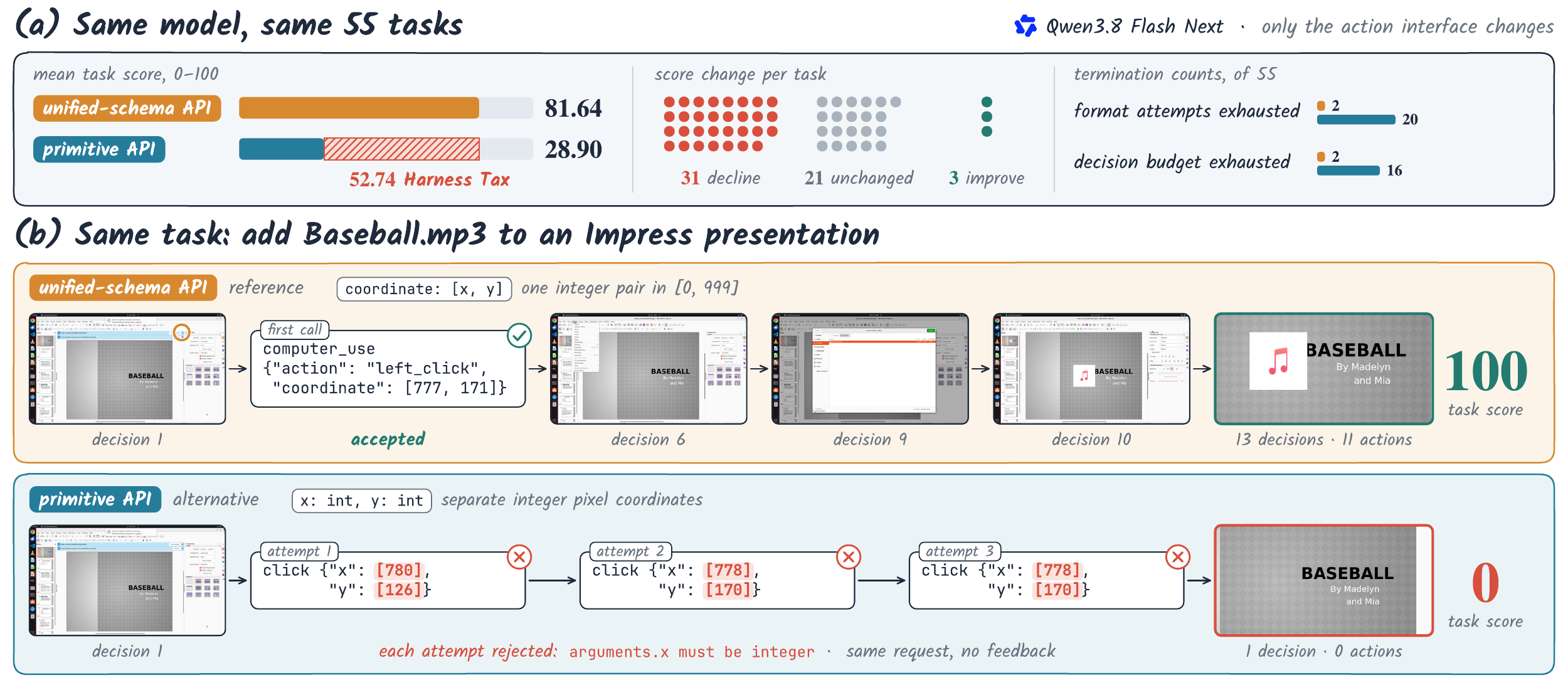}
    \caption{\textbf{Qwen3.8 Flash Next suffers from a Harness Tax when its interface changes.}
    In both panels, we fix Qwen3.8 Flash Next and compare the unified-schema API (reference, orange) with the primitive API (alternative, blue).
    (a) On the same 55 tasks, the mean task score falls from 81.64 to 28.90, a loss of 52.74 points.
    Episodes ending after repeated format errors increase from 2 to 20, and those reaching the decision budget increase from 2 to 16.
    (b) On one audio-insertion task, the score falls from 100 to zero: the reference trajectory inserts and saves the audio, whereas the alternative ends before any action is executed.
    Highlighted arrays violate the integer-coordinate requirement of the primitive API.
    Each configuration is evaluated once, and invalid responses are retried without error feedback.}
    \label{fig:harness-tax-case}
\end{figure}

Harness Tax is the drop in task performance when the same model is switched from a customized harness to an alternative one.
Fig.~\ref{fig:harness-tax-case} illustrates this effect at two levels: across a fixed set of tasks, the same model loses 52.74 points, and on a single task, it succeeds under one harness but scores zero under the other.

As shown in Sec.~\ref{sec:intro}, Qwen3.8 Flash Next exhibits a substantial performance gap between two action interfaces on 55 selected OSWorld tasks~\citep{xie2024osworld}.
Switching from the unified-schema API to the primitive API reduces the mean score from 81.64 to 28.90, a \textbf{52.74-point Harness Tax} relative to the unified-schema API (Fig.~\ref{fig:harness-tax-case}a).
The score decreases on 31 of the 55 tasks, increases on only three, and is unchanged on the remaining 21.
All other settings are held fixed: inference parameters, native tool calls, one screenshot, one recent interaction step, $1920\!\times\!1080$ input resolution, and a budget of 50 decisions.
Only the action interface changes, together with its action schemas and coordinate conventions.

The task in Fig.~\ref{fig:harness-tax-case}b asks the model to insert \texttt{Baseball.mp3} into an Impress presentation.
Under the unified-schema API, the model issues a valid \texttt{computer\_use} call, inserts the audio, and saves the file, receiving a score of 100 after 13 decisions.
Under the primitive API, \texttt{click} instead requires separate integer pixel coordinates, but the model supplies \texttt{x=[780], y=[126]}.
All three attempts violate this requirement, so the episode ends at the first decision without executing any action and receives a score of zero.
The failure thus occurs at the action interface, before the model can begin the task itself.

\section{\texorpdfstring{$Harness^{2}$}{Harness2}: A Configurable and Composable Framework}
\label{sec:framework}
\begin{figure}[t]
    \centering
    \includegraphics[width=\linewidth]{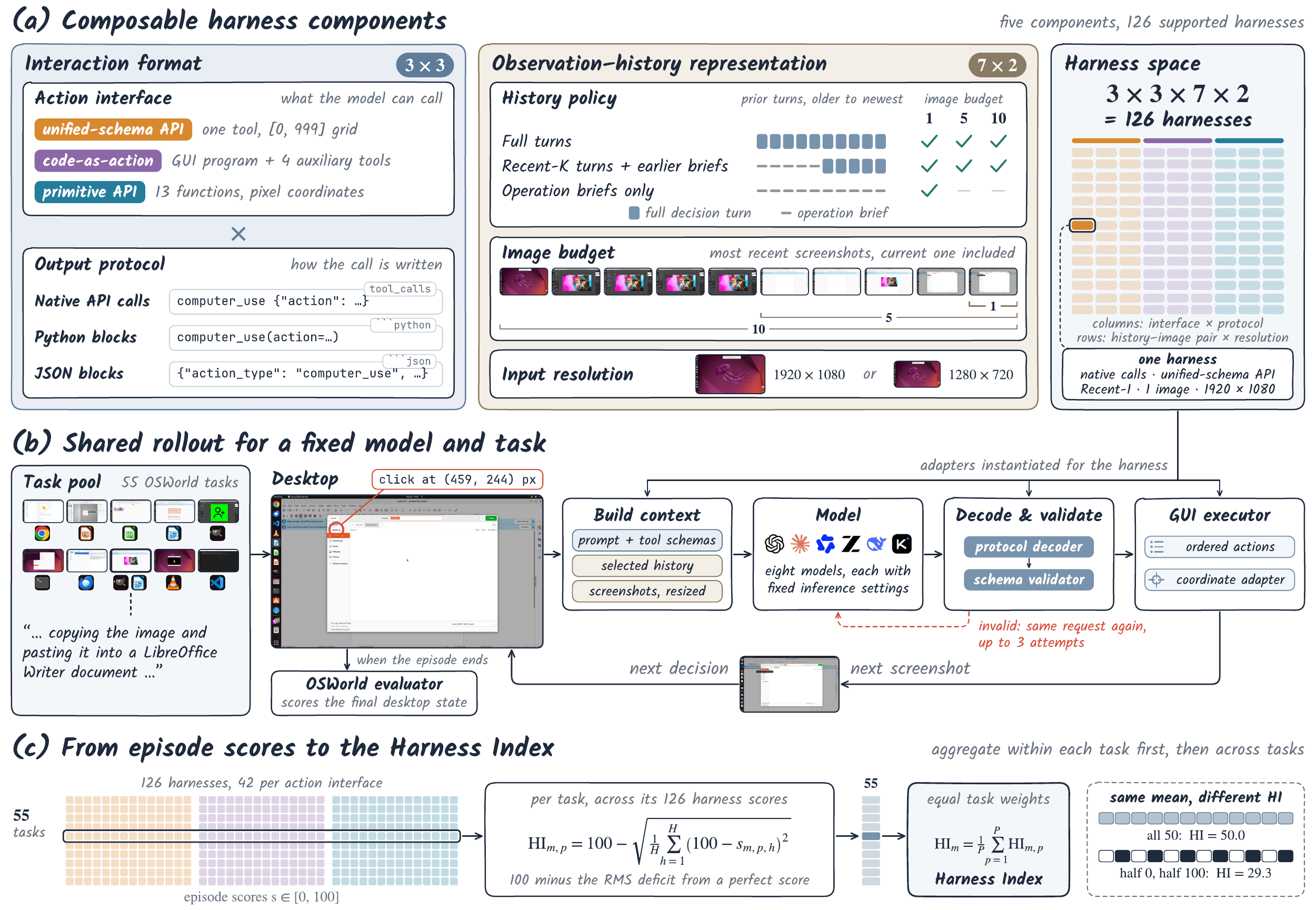}
    \caption{\textbf{The $Harness^2$ evaluation framework.}
    (a) Three action interfaces and three output protocols combine with seven valid history--image pairs and two input resolutions, yielding 126 configurations.
    Image budgets include the current observation, and recent-$K$ history summarizes earlier operations as briefs.
    (b) All configurations share the same interaction loop between the model and the desktop and differ only in how the model input is constructed and how its actions are interpreted.
    (c) Episode scores are aggregated across harnesses within each task, and the task-level HI values are then averaged across tasks.}
    \label{fig:harness2-framework}
\end{figure}

To study the interface dependence identified in Sec.~\ref{sec:harness-tax}, we design \textit{$Harness^2$}, which separates the harness components from the model, the task, and the evaluation procedure so that each component can be varied independently.
Fig.~\ref{fig:harness2-framework} summarizes the configuration space, the shared evaluation procedure, and the aggregation of scores.
We use the OSWorld tasks and their evaluation functions~\citep{xie2024osworld} and treat each combination of model, task, and harness as one evaluation unit.

\subsection{Composable Harness Components}
\label{sec:harness-components}
We organize five components into two groups: interaction format and observation--history representation.
A configuration specifies the action interface, output protocol, history policy, image budget, and input resolution.

\paragraph{Interaction format.}
The unified-schema API provides a single \texttt{computer\_use} tool whose action type is given as an argument, with coordinates normalized to integers in $[0,999]$.
The code-as-action interface accepts PyAutoGUI programs~\citep{pyautogui_docs} that may contain several actions, together with auxiliary tools for taking screenshots, waiting, and declaring completion.
The primitive API provides separate functions, such as \texttt{click}, \texttt{drag}, and \texttt{type\_text}, that take pixel coordinates as separate arguments.
Independently, the output protocol determines how actions are written in a response: as native tool calls of the model API, as function calls in a Python block, or as JSON objects in a code block.
Each protocol can be combined with each interface, giving nine interface--protocol combinations.
Python blocks are only parsed into tool calls, not run as code; under code-as-action, the PyAutoGUI program is a string argument that the tool executes.
The protocol therefore changes only the syntax of the response, independently of the action interface.
All configurations share one prompt template that describes the selected tools, the coordinate convention, and the required output format.

\paragraph{Observation--history representation.}
Full history retains all previous turns; recent-$K$ history retains the last $K$ turns and summarizes earlier ones as operation briefs; operation-brief history replaces all previous turns with such briefs.
Operation briefs are short textual descriptions constructed from the actions previously taken by the model.
The image budget limits the context to the 1, 5, or 10 most recent screenshots, including the current one; earlier turns keep their text, but their screenshots are omitted.
Full and recent-$K$ histories support all three budgets, with $K$ equal to the image budget, whereas operation-brief history uses only the current screenshot, giving seven history--image pairs.
Screenshots are provided to the model at $1920\!\times\!1080$ or $1280\!\times\!720$, while the desktop remains at $1920\!\times\!1080$; coordinates in the model's actions are rescaled to the desktop accordingly.
In total, these choices define 126 harness configurations.

\subsection{Shared Execution and Evaluation Protocol}
\label{sec:harness-protocol}
At each decision, the model receives the task instruction, the selected history, and the retained screenshots.
All actions in a response are parsed and validated before any of them is executed; after execution, the model receives a new screenshot.
A single decision may therefore contain several actions or a GUI program, and the budget of 50 decisions limits the number of model responses rather than individual actions.
If a response is invalid, the same request is sent again, up to three attempts in total, without error feedback or correction of its arguments; if all three attempts fail, the episode ends with a score of zero.
Otherwise, an episode ends when the model declares completion or reaches the decision budget, and the OSWorld evaluation function then scores the final state of the desktop.

\subsection{HI: Harness Index}
\label{sec:hi}
Let $s_{m,p,h}\in[0,100]$ be the score of model $m$ on task $p$ under harness $h$, with $H=126$ configurations.
We define task-level HI as 100 minus the root mean square deficit from a perfect score:
\begin{equation}
\begin{aligned}
\mathrm{HI}_{m,p}
&=100-\sqrt{\frac{1}{H}\sum_{h=1}^{H}(100-s_{m,p,h})^2}\\
&=100-\sqrt{(100-\mu_{m,p})^2+\sigma_{m,p}^{2}},
\end{aligned}
\label{eq:harness-index}
\end{equation}
where $\mu_{m,p}$ and $\sigma_{m,p}$ are the mean and population standard deviation of the $H$ scores.
HI lies in $[0,100]$ and equals $c$ when every harness receives the same score $c$.
At a fixed mean, greater score dispersion lowers HI, and increasing any individual score cannot lower it.
For $P$ tasks, we report
$\mathrm{HI}_{m}=\frac{1}{P}\sum_{p=1}^{P}\mathrm{HI}_{m,p},$
so that the nonlinear transformation is applied within each task and all tasks are weighted equally.
HI summarizes performance over all harnesses, whereas Harness Tax measures the loss relative to a specific reference harness.

\section{Experiments}
\label{sec:experiments}
\subsection{Experimental Setup}
\label{sec:experimental-setup}
\paragraph{Tasks and scope.}
We evaluate a fixed subset of 55 OSWorld tasks~\citep{xie2024osworld} spanning ten application domains: Chrome, GIMP, LibreOffice Calc, Impress, Writer, multi-application workflows, the operating system, Thunderbird, VLC, and VS Code.
App.~\ref{app:robustness-checks} reports further analyses of episode length and harness weighting.

\paragraph{Models and inference.}
The open-weight models are GLM 5.3 Flash~\citep{zai_glm53flash}, DeepSeek v4.1 Flash~\citep{deepseek2026v41flash}, Kimi K3 and Kimi K2.6~\citep{moonshot_kimik3,moonshot_kimik26}, and Qwen3.8 Flash Next and Qwen3.5 397B~\citep{qwen2026flashnext,qwen2026qwen35}; for Qwen3.5 397B, we use the official FP8 weights~\citep{qwen35fp8_card}.
We also evaluate the closed-source models GPT 6 Astra~\citep{openai_gpt6astra} and Claude Fable 5~\citep{anthropic2026fable5}.
Each request produces a single response of at most 16,384 output tokens, and we request the highest available reasoning effort for every model.
Overall scores reflect these inference settings as well as model capability.
For each model, the prompt, serving setup, and inference settings are identical across configurations, so that component comparisons isolate the effect of the harness.
In the main evaluation, each task is evaluated once under each configuration, with the decision budget and retry policy of Sec.~\ref{sec:harness-protocol}.

\paragraph{Scoring and comparisons.}
Mean scores and outcome rates are computed over all episodes on a 0--100 scale, including partial credit and the zero scores assigned after repeated format errors.
HI is computed from the scores of all task--harness pairs.
Paired comparisons use only tasks scored under both conditions and match all other harness components and model settings.
The image-budget and screenshot-selection studies of Qwen3.8 Flash Next in Sec.~\ref{sec:visual-history-analysis} repeat each condition three times, and we report the mean and sample standard deviation across repeats.
Budgets of 1, 5, and 10 screenshots, budgets of 2, 3, and 4 screenshots, and the two selection policies were evaluated in separate groups rather than in randomized order; we therefore interpret these comparisons, like those of the main evaluation, as descriptive.

\subsection{Performance and Robustness across Harnesses}
\label{sec:cross-harness-results}
\begin{figure}[t]
    \centering
    \includegraphics[width=0.88\linewidth]{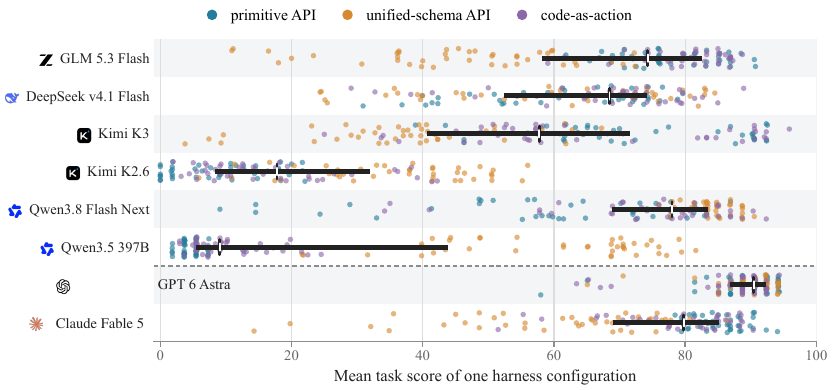}
    \caption{\textbf{Performance varies across harness configurations.}
    Each dot is one configuration's mean task score; colors identify the action interface.
    Dark horizontal bars show interquartile ranges, and vertical marks show medians.
    The dashed line separates the open-weight and closed-source models in Tab.~\ref{tab:cross-harness-results}.}
    \label{fig:score-distributions}
\end{figure}

Tab.~\ref{tab:cross-harness-results} reports mean scores, HI, and the range and spread of per-configuration mean scores, and Fig.~\ref{fig:score-distributions} shows their full distributions.
Strong performance under favorable configurations can coexist with many low-scoring configurations.
For Kimi K3, the 90th percentile of configuration means is 90.54, whereas the 10th percentile is only 33.02; for Qwen3.5 397B, the same percentiles are 70.54 and 3.64.
Over the same percentile range, GPT 6 Astra spans only 84.27--94.18, while Claude Fable 5 spans 52.45--88.33.

\begin{table}[t]
    \centering
    \caption{\textbf{Performance and robustness across harnesses.}
    Results on 55 tasks across 126 harnesses.
    In the Mean column, bold and underlined values denote the highest and second-highest means, respectively.
    The remaining columns summarize the 126 per-configuration means: worst, best, population standard deviation ($\mathrm{SD}_{\mathrm{cfg}}$), and the range from the 10th to the 90th percentile.}
    \label{tab:cross-harness-results}
    \input{tables/tab_cross_harness_results.tex}
\end{table}

The best configuration score and HI capture different aspects of performance.
Kimi K3 reaches 95.87 under its best configuration, yet its HI is only 38.17.
A high score under a single favorable harness is therefore insufficient to characterize cross-harness robustness.
Across three evaluations each of Qwen3.8 Flash Next and DeepSeek v4.1 Flash, HI varies by at most 2.75 and 1.50 points, respectively (App.~\ref{app:hi-repeatability}).

\subsection{Component Sensitivity and Interactions}
\label{sec:component-sensitivity}
\begin{figure}[t]
    \centering
    \includegraphics[width=\linewidth]{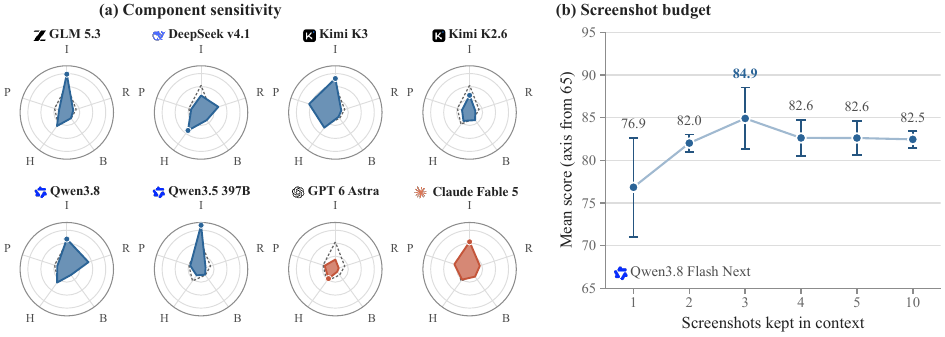}
    \caption{\textbf{Component sensitivity and screenshot budget.}
    (a)~Mean absolute score differences (pp) between configurations that differ in a single component.
    Axes: action interface (I), output protocol (P), text history (H), image budget (B), input resolution (R); rings at 10, 20, 30\,pp.
    The dot marks each model's most sensitive component; the dashed polygon is the mean over models.
    (b)~Qwen3.8 Flash Next on 55 tasks: dots show the mean across three repeats; error bars denote $\pm1$ sample SD.}
    \label{fig:component-sensitivity}
\end{figure}

To measure sensitivity to each component, we compare pairs of configurations that differ only in that component and average the absolute differences between their mean scores, weighting each pair of component settings equally (Fig.~\ref{fig:component-sensitivity}a).
Comparisons of interface, output protocol, history, and resolution use one screenshot, and comparisons of image budget use full textual history.

\paragraph{Interface dependence.}
The action interface produces the largest average absolute changes for six of the eight models.
Averaged over 18 matched settings with one screenshot, switching from the unified-schema API to the primitive API raises Kimi K3's mean score by 38.52 points but lowers that of Qwen3.8 Flash Next by 34.48 points.
The opposite preferences observed in Sec.~\ref{sec:harness-tax} therefore hold beyond a single example.

\paragraph{Protocol--interface interactions.}
The output protocol is not always a minor factor: for Kimi K3, the average absolute change in mean score caused by changing the protocol is 21.24 points.
Protocol effects also depend on the action interface.
For Qwen3.8 Flash Next, averaged over matched settings with one screenshot, switching from native calls to Python blocks lowers the mean score by 2.10 points under the unified-schema API but raises it by 14.52 points under the primitive API.
An average over interfaces can therefore obscure opposite effects of the output protocol.

\paragraph{History and visual context.}
Replacing full interaction histories with operation briefs lowers mean scores for all eight models when only one screenshot is provided.
The average losses range from 6.66 points for Qwen3.5 397B to 24.47 for DeepSeek v4.1 Flash; GPT 6 Astra also loses 11.45 points.
Resolution preferences differ as well: the lower input resolution benefits Qwen3.8 Flash Next but harms DeepSeek v4.1 Flash in matched comparisons.
These patterns call for analyzing context use separately for each model rather than assuming that a single choice of harness settings is best for all models.

\subsubsection{Visual History Budget and Selection}
\label{sec:visual-history-analysis}
With full textual history, increasing the budget from five to ten screenshots changes mean scores by $-0.94$ to $+1.88$ points across the eight models.
For DeepSeek v4.1 Flash, the average change is only 0.01 points, but the average absolute change over 18 setting pairs is 5.62, so gains and losses largely cancel.
Building on studies of positional retrieval and multimodal context use~\citep{liu2024lostmiddle,hsieh2024ruler,song2024milebench}, we further examine how the number of screenshots and the time span they cover affect performance on GUI tasks.

\paragraph{Fine-grained image budgets.}
We evaluate Qwen3.8 Flash Next with 1, 2, 3, 4, 5, and 10 screenshots, fixing the unified-schema API, Python blocks, full textual history, and $1280\!\times\!720$ input resolution.
Each budget, which includes the current screenshot, is evaluated three times on the 55 tasks.
Gains are non-monotonic: mean scores rise from 76.85 with one screenshot to 84.92 with three, then remain between 82.47 and 82.64 with four, five, and ten screenshots (Fig.~\ref{fig:component-sensitivity}b).
The highest-scoring budget varies across repeats.

\paragraph{Historical screenshot selection.}
Under the same settings, we keep five screenshots at positions $\{1,2,3,6,7\}$ or $\{1,2,3,9,10\}$, counted from the most recent.
Until ten screenshots are available, both policies keep up to the five most recent ones; the textual history and message order are unchanged.
Over three repeats, the two policies score $83.17\pm3.60$ and $85.37\pm1.28$, respectively, compared with $82.63\pm2.04$ when the five most recent screenshots are kept.
Together, these studies indicate that the benefit of visual history depends on both the number of screenshots and the time span they cover: at the same budget, combining recent and older screenshots yields higher mean scores than keeping only the most recent ones.

\subsection{Failure Mechanisms}
\label{sec:failure-mechanisms}
We distinguish failures at the action interface from failures that occur during task execution.

\paragraph{Failures at the action interface.}
In 34.38\% of Kimi K3's episodes, the model fails to produce a valid response within three attempts, which ends the episode; more than half of these failures occur at the first decision.
For Qwen3.5 397B, such failures end 82.75\% of episodes under the primitive API, compared with 18.55\% under the unified-schema API.
The invalid coordinate arguments in Sec.~\ref{sec:harness-tax} are a concrete example of such a failure.
These failures can prevent a model from carrying out a task that it completes under another interface.

\paragraph{Failures beyond output formatting.}
Episodes that reach the decision budget account for 19.80\% of all episodes for DeepSeek v4.1 Flash and 25.74\% for Qwen3.5 397B, compared with 1.27\% for Kimi K3.
Reaching the budget does not always result in a score of zero: 19.23\% of such episodes for DeepSeek v4.1 Flash still receive nonzero scores.
Thus, valid actions do not guarantee task completion, and the way an episode ends cannot by itself distinguish errors in perception, planning, or state tracking.

\paragraph{Recovery with validation feedback.}
Motivated by feedback-driven refinement~\citep{shinn2023reflexion,madaan2023selfrefine}, we additionally evaluate Qwen3.8 Flash Next on the 55 tasks and 126 configurations with validation feedback.
When a response is invalid, the rejected output and an error message describing the violation are added to the request for the next attempt, under the same validity check and limit of three attempts; invalid actions are never executed, and the feedback is removed once the decision is complete.
Among decisions whose first response is invalid, 98.70\% obtain a valid response within the three attempts.

\begin{figure}[t]
    \centering
    \includegraphics[width=\linewidth]{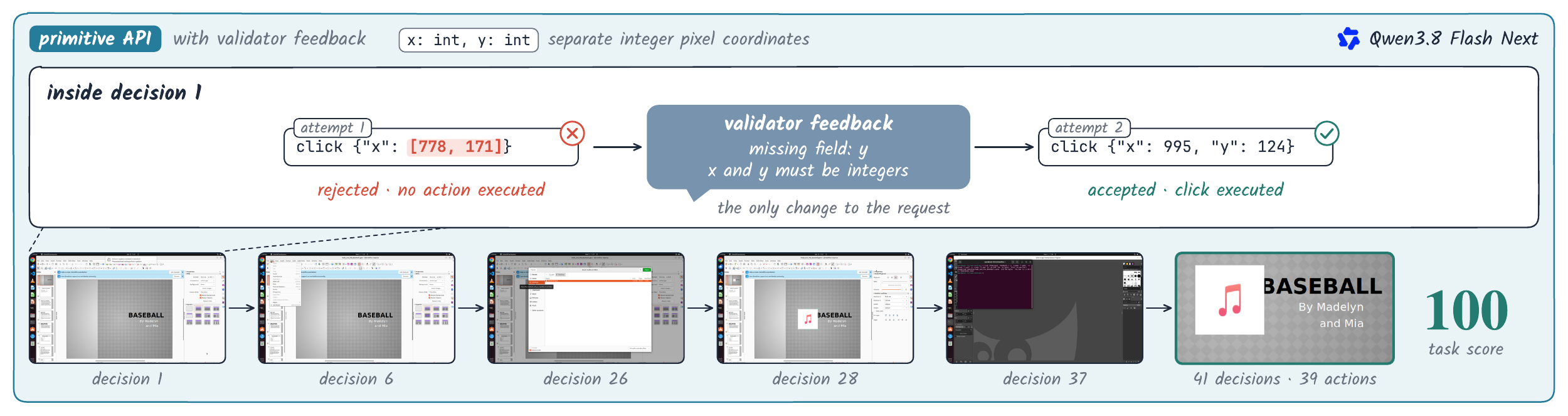}
    \caption{\textbf{Recovery from a format error on the audio-insertion task.}
    The task is the same as in Fig.~\ref{fig:harness-tax-case}, and this episode uses validation feedback.
    After a missing-field error, the model produces a valid call; the two requests differ only in the added error message.
    The final screenshot is taken at decision 41, and the task receives a score of 100.}
    \label{fig:validation-feedback-case}
\end{figure}

On the audio-insertion task of Fig.~\ref{fig:harness-tax-case}, the model with feedback initially places both coordinates in \texttt{x} and omits \texttt{y} (Fig.~\ref{fig:validation-feedback-case}).
After receiving the error message, it issues a valid click on the next attempt.
It then selects the file and inserts the audio, completing the task with a score of 100 after 41 decisions.
In this case, correcting a single format error is sufficient for the model to complete the task.

\textbf{Format recovery does not guarantee task completion.}
Among episodes with at least one such recovery, 54.27\% receive full credit, 7.80\% partial credit, and 37.93\% zero.
In the three configurations shared with the image-budget study in Sec.~\ref{sec:visual-history-analysis}, the mean score is also higher with feedback (App.~\ref{app:feedback-comparison}).

\subsection{SFT for Cross-Harness Robustness}
\label{sec:sft-robustness}
We further examine whether supervised fine-tuning (SFT) improves the performance of Qwen3.8 27B across harnesses.
The training data contain approximately 300K step-level samples, of which about 10\% span multiple harnesses, and the model is evaluated on the same 55 tasks and 126 configurations.
SFT raises the mean score from 63.19 to 68.30 and HI from 42.78 to 46.86, and it increases the mean score for every setting of each component.
Training and evaluation details are given in App.~\ref{app:sft-details}.

\section{Conclusion and Future Work}
In this work, we identify Harness Tax and show that changing the harness can substantially degrade performance and reverse the rankings of the same models.
To characterize this dependence systematically, we introduce \textit{$Harness^2$} with 126 composable harness configurations and propose the Harness Index (HI) to jointly assess task performance and cross-harness stability.
The benchmark reveals model-specific interface preferences, history sensitivity, and distinct patterns of format and decision-budget exhaustion.
We also observe format recovery with validation feedback and broad task-score gains after supervised fine-tuning.

We would like to highlight two directions for future research:
\begin{itemize}
    \item \textbf{Generalization to Unseen Harnesses.}
    An important next step is to extend gains from fine-tuning to unseen action interfaces, component combinations, and tasks, building on cross-platform GUI learning~\citep{xu2024aguvis}.

    \item \textbf{Test-Time Adaptation to Harnesses.}
    The observed recovery with validation feedback motivates adaptation mechanisms that use interface specifications and interaction feedback to preserve effective task execution under unfamiliar harnesses~\citep{shinn2023reflexion,agashe2025agents2,agashe2026evolving}.
\end{itemize}
We hope this work motivates the community to pay more attention to the harness tax issue and develop harness-agnostic GUI agents whose capabilities remain reliable across deployment settings.

\subsection*{AI use statement}
An AI assistant supported literature organization, manuscript drafting and editing, and the implementation of analysis scripts and visualizations for existing evaluation records. The authors are responsible for reviewing the AI-assisted material and for the final claims, analyses, and conclusions.

\clearpage
\bibliographystyle{plainnat}
\bibliography{cite}

\clearpage
\appendix
\input{appendix_robustness_checks}
\input{appendix_action_interfaces}

\end{document}

%% file: math_commands.tex
\usepackage{amsmath,amsfonts,bm}

\def\eqref#1{equation~\ref{#1}}

\def\1{\bm{1}}

\DeclareMathAlphabet{\mathsfit}{\encodingdefault}{\sfdefault}{m}{sl}
\SetMathAlphabet{\mathsfit}{bold}{\encodingdefault}{\sfdefault}{bx}{n}



%% file: tables/tab_cross_harness_results.tex
\ifdim\linewidth<300pt
{\fontsize{8}{9.5}\selectfont
\setlength{\tabcolsep}{1pt}
\renewcommand{\arraystretch}{1.12}
\begin{tabular}{@{}l!{\hspace{1.5pt}}cc!{\hspace{2pt}}cccc}
\toprule
 & \multicolumn{2}{c}{Overall} & \multicolumn{4}{c}{Harness configurations} \\
\cmidrule(lr){2-3}\cmidrule(lr){4-7}
Model & Mean & HI & Worst & Best & $\mathrm{SD}_{\mathrm{cfg}}$ & P10--P90 \\
\midrule
\multicolumn{7}{@{}l}{\textit{Open-weight models}} \\
\raisebox{-0.22em}{\includegraphics[height=0.95em]{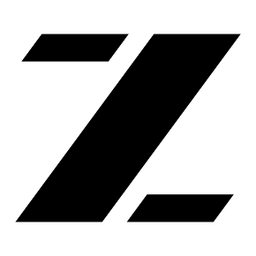}}\hspace{0.25em}GLM 5.3 Flash & \cellcolor[HTML]{C7D6E4}67.99 & \cellcolor[HTML]{D1DDE8}46.60 & \cellcolor[HTML]{EAF4F3}10.91 & \cellcolor[HTML]{BBDCD8}90.71 & \cellcolor[HTML]{F1D6D0}19.09 & \cellcolor[HTML]{CDE5E3}38.91--86.70 \\
\arrayrulecolor[gray]{0.78}\specialrule{0.25pt}{0pt}{0pt}\arrayrulecolor{black}
\raisebox{-0.22em}{\includegraphics[height=0.95em]{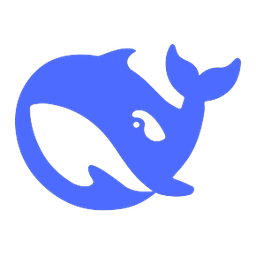}}\hspace{0.25em}DeepSeek v4.1 Flash & \cellcolor[HTML]{CCDAE6}62.87 & \cellcolor[HTML]{D4DFEA}43.73 & \cellcolor[HTML]{DAECEA}24.35 & \cellcolor[HTML]{BEDEDA}88.91 & \cellcolor[HTML]{F4E0DB}15.58 & \cellcolor[HTML]{D1E7E5}39.43--79.56 \\
\arrayrulecolor[gray]{0.78}\specialrule{0.25pt}{0pt}{0pt}\arrayrulecolor{black}
\raisebox{-0.22em}{\includegraphics[height=0.95em]{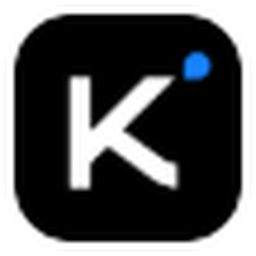}}\hspace{0.25em}Kimi K3 & \cellcolor[HTML]{D1DDE8}58.27 & \cellcolor[HTML]{DAE4ED}38.17 & \cellcolor[HTML]{F2F8F8}3.77 & \cellcolor[HTML]{B2D7D3}95.87 & \cellcolor[HTML]{EECEC7}21.90 & \cellcolor[HTML]{CEE6E3}33.02--90.54 \\
\arrayrulecolor[gray]{0.78}\specialrule{0.25pt}{0pt}{0pt}\arrayrulecolor{black}
\raisebox{-0.22em}{\includegraphics[height=0.95em]{figure/component_scores/model_icons/kimi.png}}\hspace{0.25em}Kimi K2.6 & \cellcolor[HTML]{F7F9FB}20.45 & \cellcolor[HTML]{F7F9FB}12.31 & \cellcolor[HTML]{F6FBFA}0.00 & \cellcolor[HTML]{F6FBFA}56.00 & \cellcolor[HTML]{F5E1DC}15.21 & \cellcolor[HTML]{F6FBFA}1.87--43.58 \\
\arrayrulecolor[gray]{0.78}\specialrule{0.25pt}{0pt}{0pt}\arrayrulecolor{black}
\raisebox{-0.22em}{\includegraphics[height=0.95em]{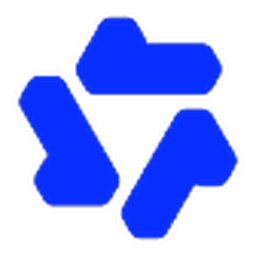}}\hspace{0.25em}Qwen3.8 Flash Next & \cellcolor[HTML]{C2D2E1}72.89 & \cellcolor[HTML]{CAD8E5}52.39 & \cellcolor[HTML]{ECF5F4}9.09 & \cellcolor[HTML]{B8DBD7}92.36 & \cellcolor[HTML]{F3DCD7}16.87 & \cellcolor[HTML]{C8E2E0}49.80--86.90 \\
\arrayrulecolor[gray]{0.78}\specialrule{0.25pt}{0pt}{0pt}\arrayrulecolor{black}
\raisebox{-0.22em}{\includegraphics[height=0.95em]{figure/component_scores/model_icons/qwen.png}}\hspace{0.25em}Qwen3.5 397B & \cellcolor[HTML]{F2F6F9}24.62 & \cellcolor[HTML]{F4F7FA}14.48 & \cellcolor[HTML]{F4F9F9}1.82 & \cellcolor[HTML]{CBE4E1}81.62 & \cellcolor[HTML]{EAC2B8}26.40 & \cellcolor[HTML]{E8F3F2}3.64--70.54 \\
\midrule
\multicolumn{7}{@{}l}{\textit{Closed-source models}} \\
\raisebox{-0.22em}{\includegraphics[height=0.95em]{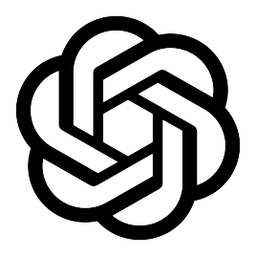}}\hspace{0.25em}GPT 6 Astra & \cellcolor[HTML]{B3C7DA}\textbf{88.31} & \cellcolor[HTML]{B3C7DA}73.62 & \cellcolor[HTML]{B2D7D3}57.99 & \cellcolor[HTML]{B5D9D5}94.37 & \cellcolor[HTML]{FDF8F7}6.78 & \cellcolor[HTML]{B2D7D3}84.27--94.18 \\
\arrayrulecolor[gray]{0.78}\specialrule{0.25pt}{0pt}{0pt}\arrayrulecolor{black}
\raisebox{-0.22em}{\includegraphics[height=0.95em]{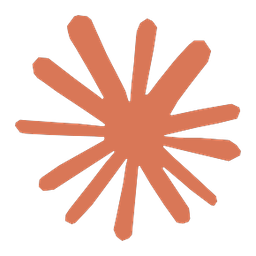}}\hspace{0.25em}Claude Fable 5 & \cellcolor[HTML]{BFD1E0}\underline{75.58} & \cellcolor[HTML]{C5D4E2}57.28 & \cellcolor[HTML]{E6F2F1}14.29 & \cellcolor[HTML]{B5D9D5}94.07 & \cellcolor[HTML]{F5E1DC}15.17 & \cellcolor[HTML]{C6E1DE}52.45--88.33 \\
\bottomrule
\end{tabular}
}
\else
{\small
\setlength{\tabcolsep}{5.5pt}
\renewcommand{\arraystretch}{1.12}
\begin{tabular}{@{}lcc!{\hspace{4pt}}cccc}
\toprule
 & \multicolumn{2}{c}{Overall} & \multicolumn{4}{c}{Across harness configurations} \\
\cmidrule(lr){2-3}\cmidrule(lr){4-7}
Model & Mean & HI & Worst & Best & $\mathrm{SD}_{\mathrm{cfg}}$ & P10--P90 \\
\midrule
\multicolumn{7}{@{}l}{\textit{Open-weight models}} \\
\hspace{0.6em}\raisebox{-0.22em}{\includegraphics[height=0.95em]{figure/component_scores/model_icons/glm.png}}\hspace{0.35em}GLM 5.3 Flash & \cellcolor[HTML]{C7D6E4}67.99 & \cellcolor[HTML]{D1DDE8}46.60 & \cellcolor[HTML]{EAF4F3}10.91 & \cellcolor[HTML]{BBDCD8}90.71 & \cellcolor[HTML]{F1D6D0}19.09 & \cellcolor[HTML]{CDE5E3}38.91--86.70 \\
\hspace{0.6em}\raisebox{-0.22em}{\includegraphics[height=0.95em]{figure/component_scores/model_icons/deepseek.png}}\hspace{0.35em}DeepSeek v4.1 Flash & \cellcolor[HTML]{CCDAE6}62.87 & \cellcolor[HTML]{D4DFEA}43.73 & \cellcolor[HTML]{DAECEA}24.35 & \cellcolor[HTML]{BEDEDA}88.91 & \cellcolor[HTML]{F4E0DB}15.58 & \cellcolor[HTML]{D1E7E5}39.43--79.56 \\
\hspace{0.6em}\raisebox{-0.22em}{\includegraphics[height=0.95em]{figure/component_scores/model_icons/kimi.png}}\hspace{0.35em}Kimi K3 & \cellcolor[HTML]{D1DDE8}58.27 & \cellcolor[HTML]{DAE4ED}38.17 & \cellcolor[HTML]{F2F8F8}3.77 & \cellcolor[HTML]{B2D7D3}95.87 & \cellcolor[HTML]{EECEC7}21.90 & \cellcolor[HTML]{CEE6E3}33.02--90.54 \\
\hspace{0.6em}\raisebox{-0.22em}{\includegraphics[height=0.95em]{figure/component_scores/model_icons/kimi.png}}\hspace{0.35em}Kimi K2.6 & \cellcolor[HTML]{F7F9FB}20.45 & \cellcolor[HTML]{F7F9FB}12.31 & \cellcolor[HTML]{F6FBFA}0.00 & \cellcolor[HTML]{F6FBFA}56.00 & \cellcolor[HTML]{F5E1DC}15.21 & \cellcolor[HTML]{F6FBFA}1.87--43.58 \\
\hspace{0.6em}\raisebox{-0.22em}{\includegraphics[height=0.95em]{figure/component_scores/model_icons/qwen.png}}\hspace{0.35em}Qwen3.8 Flash Next & \cellcolor[HTML]{C2D2E1}72.89 & \cellcolor[HTML]{CAD8E5}52.39 & \cellcolor[HTML]{ECF5F4}9.09 & \cellcolor[HTML]{B8DBD7}92.36 & \cellcolor[HTML]{F3DCD7}16.87 & \cellcolor[HTML]{C8E2E0}49.80--86.90 \\
\hspace{0.6em}\raisebox{-0.22em}{\includegraphics[height=0.95em]{figure/component_scores/model_icons/qwen.png}}\hspace{0.35em}Qwen3.5 397B & \cellcolor[HTML]{F2F6F9}24.62 & \cellcolor[HTML]{F4F7FA}14.48 & \cellcolor[HTML]{F4F9F9}1.82 & \cellcolor[HTML]{CBE4E1}81.62 & \cellcolor[HTML]{EAC2B8}26.40 & \cellcolor[HTML]{E8F3F2}3.64--70.54 \\
\midrule
\multicolumn{7}{@{}l}{\textit{Closed-source models}} \\
\hspace{0.6em}\raisebox{-0.22em}{\includegraphics[height=0.95em]{figure/component_scores/model_icons/openai.png}}\hspace{0.35em}GPT 6 Astra & \cellcolor[HTML]{B3C7DA}\textbf{88.31} & \cellcolor[HTML]{B3C7DA}73.62 & \cellcolor[HTML]{B2D7D3}57.99 & \cellcolor[HTML]{B5D9D5}94.37 & \cellcolor[HTML]{FDF8F7}6.78 & \cellcolor[HTML]{B2D7D3}84.27--94.18 \\
\hspace{0.6em}\raisebox{-0.22em}{\includegraphics[height=0.95em]{figure/component_scores/model_icons/claude.png}}\hspace{0.35em}Claude Fable 5 & \cellcolor[HTML]{BFD1E0}\underline{75.58} & \cellcolor[HTML]{C5D4E2}57.28 & \cellcolor[HTML]{E6F2F1}14.29 & \cellcolor[HTML]{B5D9D5}94.07 & \cellcolor[HTML]{F5E1DC}15.17 & \cellcolor[HTML]{C6E1DE}52.45--88.33 \\
\bottomrule
\end{tabular}
}
\fi

%% file: appendix_robustness_checks.tex
\providecommand{\supponlycite}[1]{}
\section{Supplementary Analyses}
\label{app:robustness-checks}

Unless stated otherwise, the analyses below use the 55 OSWorld tasks~\citep{xie2024osworld}, the 126 harness configurations, and the eight models of the main study: GLM 5.3 Flash~\citep{zai_glm53flash}, DeepSeek v4.1 Flash~\citep{deepseek2026v41flash}, Kimi K3 and Kimi K2.6~\citep{moonshot_kimik3,moonshot_kimik26}, Qwen3.8 Flash Next and Qwen3.5 397B~\citep{qwen2026flashnext,qwen2026qwen35}, GPT 6 Astra~\citep{openai_gpt6astra}, and Claude Fable 5~\citep{anthropic2026fable5}.

\subsection{Action Interfaces and Rollout Length}
Changing the action interface changes the entire action space presented to the model.
The primitive API uses separate typed functions and image-pixel coordinates; the unified-schema API uses one action-discriminated schema with normalized coordinates; code-as-action accepts executable PyAutoGUI~\citep{pyautogui_docs} programs whose coordinates are given in screenshot pixels.
Under every output protocol, a single response may contain several tool calls, so no interface is limited to one action per decision.
Code-as-action further supports loops and provides tools for taking screenshots and waiting.
Because the interfaces differ in several respects at once, a comparison between them reflects each interface as a whole rather than the effect of any single design choice.

Tab.~\ref{tab:rollout-lengths} reports the mean number of decisions per episode.
The table describes episode length rather than computational cost; fewer decisions may also indicate early failure.

\begin{table}[!h]
\centering\small
\caption{Average number of decisions per episode under each action interface, computed over all tasks and harness configurations that use the interface. A decision is one model response, which may contain several tool calls.}
\label{tab:rollout-lengths}
\input{tables/tab_rollout_lengths.tex}
\end{table}

The action interface substantially changes episode length, but not in a consistent direction: for six of the eight models, the longest average across the three interfaces is at least 1.5 times the shortest, and the unified-schema API lengthens episodes relative to the primitive API for five models, most markedly for Claude Fable 5 (16.43 vs.\ 8.29) and Kimi K2.6 (23.06 vs.\ 12.54), but shortens them for the other three.
Code-as-action yields the shortest episodes for GLM 5.3 Flash, DeepSeek v4.1 Flash, Kimi K3, and GPT 6 Astra, but the longest for Kimi K2.6 and Qwen3.5 397B (36.06 and 37.86), 2.9 and 4.6 times their primitive-API values.
The two closed-source models stay below 17 decisions under every interface, whereas the open-weight models vary considerably more from one interface to another, with the longest and shortest averages of a single model differing by up to a factor of 4.6, compared with at most 2.0 for the closed-source models.

\subsection{Comparison with Validation Feedback}
\label{app:feedback-comparison}
To examine whether validation feedback also affects task scores, we compare two evaluations of Qwen3.8 Flash Next from the main paper: the evaluation with validation feedback and the image-budget study, which uses no feedback.
The two evaluations share three harness configurations, all with the unified-schema API, Python blocks, full textual history, and $1280\!\times\!720$ resolution, and differing only in the number of screenshots (one, five, or ten).
Apart from the feedback itself, both evaluations use the same tasks, prompts, inference parameters, and environment.
Tab.~\ref{tab:existing-feedback-comparison} shows that the mean score is higher with feedback in all three configurations, by 1.46 to 4.40 points.

\begin{table}[!htb]
\centering\small
\caption{Mean scores with and without validation feedback for one, five, or ten screenshots, with all other harness settings held fixed.}
\label{tab:existing-feedback-comparison}
\input{tables/tab_existing_feedback_comparison.tex}
\end{table}

The gain from feedback grows with the number of screenshots, from $+$1.46 points with one screenshot to $+$2.60 with five and $+$4.40 with ten.
Without feedback, the mean score saturates beyond five screenshots (82.40 and 82.21), whereas with feedback it continues to rise from 85.00 to 86.61.
With one screenshot, the gain is small relative to the variation across the three evaluations without feedback, whose means range from 70.47 to 83.02; with five or ten screenshots, the score with feedback exceeds that of every evaluation without feedback.
These results suggest that validation feedback is most useful when the model can also see more of its recent interaction.

\subsection{Repeatability of HI}
\label{app:hi-repeatability}
We evaluate Qwen3.8 Flash Next and DeepSeek v4.1 Flash three times (Repeats 1--3) on the same 55 tasks and 126 harness configurations.
Because a complete evaluation is costly, repeated evaluations are limited to these two models.
All three evaluations are scored with the same rule, which is also used to compute mean task scores.

\begin{table}[!htb]
\centering
\caption{Mean and HI over three evaluations scored with the same rule. HI is computed within each task and then averaged across tasks for each evaluation.}
\label{tab:hi-repeatability}
\input{tables/tab_hi_repeatability.tex}
\end{table}

Across the three evaluations, HI varies by at most 2.75 points for Qwen3.8 Flash Next and 1.50 points for DeepSeek v4.1 Flash, and Qwen3.8 Flash Next has the higher HI in every evaluation.
The mean score varies by a similar amount, at most 2.59 and 1.50 points, respectively.
Both variations are small compared with the HI gap between the two models, which ranges from 7.38 to 8.66 points across evaluations, so the comparison between them does not depend on which evaluation is used.

\subsection{Evaluation of the Fine-tuned Model}
\label{app:sft-details}
We evaluate Qwen3.8 27B~\citep{qwen38} before and after SFT on the same 55 OSWorld tasks~\citep{xie2024osworld} and 126 harness configurations as in the main study.

\paragraph{Training data.}
The SFT data contain approximately 300K step-level samples, each corresponding to a single decision step.
The training tasks are drawn from ProCUA-SFT~\citep{jung2026procua} and from a dataset that we curated, and none of them is among the 55 evaluation tasks.
About 90\% of the samples are collected under a single harness, and the remaining 10\% are cross-harness data collected under all 126 harness configurations.

\paragraph{Training setup.}
We fine-tune all parameters of Qwen3.8 27B, including both the language model and the vision encoder, for one epoch.
Optimization uses AdamW~\citep{loshchilov2019adamw} with $\beta_1=0.9$, $\beta_2=0.999$, and $\epsilon=10^{-8}$, a constant learning rate of $2\times10^{-6}$, no weight decay, and gradient clipping at a maximum norm of 1.0.
The base and fine-tuned models are evaluated with the same inference settings, listed in Sec.~\ref{app:eval-setup}.

\paragraph{Overall results.}
Tab.~\ref{tab:sft-summary} shows that SFT raises the mean from 63.19 to 68.30 ($+$5.11) and HI from 42.78 to 46.86 ($+$4.08).
The configuration-level mean increases for 95 of the 126 configurations.

\begin{table}[!htb]
\centering\small
\caption{Qwen3.8 27B before (Base) and after SFT on 55 tasks and 126 configurations (0--100 scale). Change is SFT minus Base.}
\label{tab:sft-summary}
\input{analysis_data/sft/tab_sft_summary.tex}
\end{table}

\paragraph{Results by component.}
Tab.~\ref{tab:sft-components} reports the mean score for each component setting before and after SFT.
The mean increases for every setting, but the size of the increase varies.
Among output protocols, native tool calls gain the most ($+$9.50), compared with $+$2.11 for Python blocks and $+$3.71 for JSON blocks.
Among action interfaces, the primitive API gains $+$7.20, more than the unified-schema API ($+$5.16) and code-as-action ($+$3.02), although it remains the lowest-scoring interface.
Gains are larger with five or ten screenshots ($+$6.79 and $+$8.75) than with one ($+$1.51), and operation briefs change little ($+$0.46).

\begin{table}[!htb]
\centering\small
\caption{Mean scores of Qwen3.8 27B by component setting before (Base) and after SFT. Change is SFT minus Base, and the bars show the change on a scale from 0 to $+$10 points.}
\label{tab:sft-components}
\input{analysis_data/sft/tab_sft_components.tex}
\end{table}

Although cross-harness data make up only about 10\% of the training set, the fine-tuned model improves on 95 of the 126 configurations and on every component setting.
This suggests that a small fraction of cross-harness data can substantially improve generalization across harnesses, pointing to a data-efficient route to cross-harness robustness.

\subsection{Sensitivity to Harness Weights}
\label{app:harness-weight-sensitivity}
The main HI assigns equal weight to each of the 126 configurations, which assigns total weights $3/7$, $3/7$, and $1/7$ to full, recent-$K$, and operation-brief histories.
To assess sensitivity to this evaluation distribution, we also assign one third of the weight to each history policy and distribute that weight uniformly within its configurations.
For normalized weights $\sum_h w_h=1$, the corresponding task-level index is
\begin{equation}
\mathrm{HI}_{m,p}^{(w)}=100-\sqrt{\sum_h w_h(100-s_{m,p,h})^2}.
\end{equation}
As in the main study, we then average over the $P$ tasks, $\mathrm{HI}_{m}^{(w)}=\frac{1}{P}\sum_{p=1}^{P}\mathrm{HI}_{m,p}^{(w)}$.

Tab.~\ref{tab:hi-weight-check} shows that absolute HI values change under this reweighting, while the model ordering is preserved.
Balancing histories also changes the marginal screenshot-budget distribution because only one screenshot is supported for operation briefs.

\begin{table}[!htb]
\centering\small
\caption{HI under uniform configuration weights and equal total weight for each history policy. HI change is the latter minus the former; darker cells mark larger reductions.}
\label{tab:hi-weight-check}
\input{tables/tab_hi_weight_check.tex}
\end{table}

\subsection{Evaluation Tasks and Inference Settings}
\label{app:eval-setup}
The 55 evaluation tasks are drawn from OSWorld~\citep{xie2024osworld} and cover ten application domains with three to six tasks each (Tab.~\ref{tab:task-domains}).
Tab.~\ref{tab:inference-settings} lists the inference settings of each model.
These settings are specified in each request and are identical across all harness configurations.
We request a reasoning effort of xhigh for every model except Qwen3.5 397B, for which we request high, the highest level accepted by its serving interface.
Effort levels are defined separately by each model interface and do not necessarily correspond to the same amount of reasoning across models.
The open-weight models use a temperature of 0.6; for GPT 6 Astra and Claude Fable 5, the temperature is left at the service default.

\begin{table}[!htb]
\centering\small
\begin{minipage}[t]{0.34\linewidth}
\centering
\caption{Evaluation tasks per application domain.}
\label{tab:task-domains}
\setlength{\tabcolsep}{9pt}
\renewcommand{\arraystretch}{1.12}
\begin{tabular}{@{}lc@{}}
\toprule
Domain & Tasks \\
\midrule
Chrome & 3 \\
GIMP & 6 \\
\multicolumn{2}{@{}l}{\textit{LibreOffice}} \\
\hspace{0.6em}Calc & 6 \\
\hspace{0.6em}Impress & 6 \\
\hspace{0.6em}Writer & 6 \\
Multi-application & 5 \\
Operating system & 6 \\
Thunderbird & 6 \\
VLC & 5 \\
VS Code & 6 \\
\midrule
Total & 55 \\
\bottomrule
\end{tabular}
\end{minipage}\hfill
\begin{minipage}[t]{0.60\linewidth}
\centering
\caption{Inference settings specified in each request. A dash indicates that the parameter is not specified and the service default applies.}
\label{tab:inference-settings}
\setlength{\tabcolsep}{9pt}
\renewcommand{\arraystretch}{1.12}
\begin{tabular}{@{}lcc@{}}
\toprule
Model & Reasoning effort & Temperature \\
\midrule
\multicolumn{3}{@{}l}{\textit{Open-weight models}} \\
\hspace{0.6em}\raisebox{-0.22em}{\includegraphics[height=0.95em]{figure/component_scores/model_icons/glm.png}}\hspace{0.35em}GLM 5.3 Flash & xhigh & 0.6 \\
\hspace{0.6em}\raisebox{-0.22em}{\includegraphics[height=0.95em]{figure/component_scores/model_icons/deepseek.png}}\hspace{0.35em}DeepSeek v4.1 Flash & xhigh & 0.6 \\
\hspace{0.6em}\raisebox{-0.22em}{\includegraphics[height=0.95em]{figure/component_scores/model_icons/kimi.png}}\hspace{0.35em}Kimi K3 & xhigh & 0.6 \\
\hspace{0.6em}\raisebox{-0.22em}{\includegraphics[height=0.95em]{figure/component_scores/model_icons/kimi.png}}\hspace{0.35em}Kimi K2.6 & xhigh & 0.6 \\
\hspace{0.6em}\raisebox{-0.22em}{\includegraphics[height=0.95em]{figure/component_scores/model_icons/qwen.png}}\hspace{0.35em}Qwen3.8 Flash Next & xhigh & 0.6 \\
\hspace{0.6em}\raisebox{-0.22em}{\includegraphics[height=0.95em]{figure/component_scores/model_icons/qwen.png}}\hspace{0.35em}Qwen3.5 397B & high & 0.6 \\
\midrule
\multicolumn{3}{@{}l}{\textit{Closed-source models}} \\
\hspace{0.6em}\raisebox{-0.22em}{\includegraphics[height=0.95em]{figure/component_scores/model_icons/openai.png}}\hspace{0.35em}GPT 6 Astra & xhigh & -- \\
\hspace{0.6em}\raisebox{-0.22em}{\includegraphics[height=0.95em]{figure/component_scores/model_icons/claude.png}}\hspace{0.35em}Claude Fable 5 & xhigh & -- \\
\midrule
\multicolumn{3}{@{}l}{\textit{Fine-tuned model}} \\
\hspace{0.6em}\raisebox{-0.22em}{\includegraphics[height=0.95em]{figure/component_scores/model_icons/qwen.png}}\hspace{0.35em}Qwen3.8 27B (Base, SFT) & xhigh & 0.6 \\
\bottomrule
\end{tabular}
\end{minipage}
\end{table}

%% file: tables/tab_rollout_lengths.tex
{\small
\setlength{\tabcolsep}{10pt}
\renewcommand{\arraystretch}{1.12}
\begin{tabular}{@{}lccc}
\toprule
Model & Primitive & Unified-schema & Code-as-action \\
\midrule
\multicolumn{4}{@{}l}{\textit{Open-weight models}} \\
\hspace{0.6em}\raisebox{-0.22em}{\includegraphics[height=0.95em]{figure/component_scores/model_icons/glm.png}}\hspace{0.35em}GLM 5.3 Flash & \cellcolor[HTML]{EEF5F7}10.86 & \cellcolor[HTML]{FAEFE3}16.27 & \cellcolor[HTML]{F7F4F9}10.05 \\
\hspace{0.6em}\raisebox{-0.22em}{\includegraphics[height=0.95em]{figure/component_scores/model_icons/deepseek.png}}\hspace{0.35em}DeepSeek v4.1 Flash & \cellcolor[HTML]{CFE2E9}24.61 & \cellcolor[HTML]{F7E7D4}22.97 & \cellcolor[HTML]{EAE4EF}20.67 \\
\hspace{0.6em}\raisebox{-0.22em}{\includegraphics[height=0.95em]{figure/component_scores/model_icons/kimi.png}}\hspace{0.35em}Kimi K3 & \cellcolor[HTML]{F4F8FA}8.06 & \cellcolor[HTML]{FCF5EE}11.15 & \cellcolor[HTML]{FAF9FC}6.99 \\
\hspace{0.6em}\raisebox{-0.22em}{\includegraphics[height=0.95em]{figure/component_scores/model_icons/kimi.png}}\hspace{0.35em}Kimi K2.6 & \cellcolor[HTML]{EAF2F5}12.54 & \cellcolor[HTML]{F7E7D4}23.06 & \cellcolor[HTML]{D8CBE1}36.06 \\
\hspace{0.6em}\raisebox{-0.22em}{\includegraphics[height=0.95em]{figure/component_scores/model_icons/qwen.png}}\hspace{0.35em}Qwen3.8 Flash Next & \cellcolor[HTML]{D4E5EB}22.28 & \cellcolor[HTML]{FBF3EA}13.07 & \cellcolor[HTML]{EAE3EF}21.13 \\
\hspace{0.6em}\raisebox{-0.22em}{\includegraphics[height=0.95em]{figure/component_scores/model_icons/qwen.png}}\hspace{0.35em}Qwen3.5 397B & \cellcolor[HTML]{F3F8FA}8.31 & \cellcolor[HTML]{FBF3EA}12.93 & \cellcolor[HTML]{D6C9E0}37.86 \\
\midrule
\multicolumn{4}{@{}l}{\textit{Closed-source models}} \\
\hspace{0.6em}\raisebox{-0.22em}{\includegraphics[height=0.95em]{figure/component_scores/model_icons/openai.png}}\hspace{0.35em}GPT 6 Astra & \cellcolor[HTML]{EFF5F7}10.46 & \cellcolor[HTML]{FCF6F0}10.33 & \cellcolor[HTML]{F8F6FA}8.79 \\
\hspace{0.6em}\raisebox{-0.22em}{\includegraphics[height=0.95em]{figure/component_scores/model_icons/claude.png}}\hspace{0.35em}Claude Fable 5 & \cellcolor[HTML]{F3F8FA}8.29 & \cellcolor[HTML]{FAEFE2}16.43 & \cellcolor[HTML]{F3F0F6}12.86 \\
\bottomrule
\end{tabular}
}

%% file: tables/tab_existing_feedback_comparison.tex
{\small
\setlength{\tabcolsep}{10pt}
\renewcommand{\arraystretch}{1.12}
\begin{tabular}{@{}lcc!{\hspace{4pt}}c}
\toprule
Screenshots & No feedback & Validation feedback & Difference \\
\midrule
1 & \cellcolor[HTML]{F9F9FA}76.19 & \cellcolor[HTML]{F6FBFA}77.66 & \cellcolor[HTML]{F6FAF9}+1.46 \\
5 & \cellcolor[HTML]{C5CDD4}82.40 & \cellcolor[HTML]{BFDEDA}85.00 & \cellcolor[HTML]{DCECE7}+2.60 \\
10 & \cellcolor[HTML]{C7CED5}82.21 & \cellcolor[HTML]{B2D7D3}86.61 & \cellcolor[HTML]{B2D6CB}+4.40 \\
\bottomrule
\end{tabular}
}

%% file: tables/tab_hi_repeatability.tex
{\small
\setlength{\tabcolsep}{5.5pt}
\renewcommand{\arraystretch}{1.12}
\begin{tabular}{@{}lccc!{\hspace{4pt}}ccc@{}}
\toprule
 & \multicolumn{3}{c}{Mean} & \multicolumn{3}{c}{HI} \\
\cmidrule(lr){2-4}\cmidrule(lr){5-7}
Model & Repeat 1 & Repeat 2 & Repeat 3 & Repeat 1 & Repeat 2 & Repeat 3 \\
\midrule
Qwen3.8 Flash Next & 73.11 & 70.78 & 70.51 & \cellcolor[HTML]{EAF0F6}52.39 & \cellcolor[HTML]{EAF0F6}49.73 & \cellcolor[HTML]{EAF0F6}49.64 \\
DeepSeek v4.1 Flash & 63.42 & 61.97 & 61.92 & \cellcolor[HTML]{EAF0F6}43.73 & \cellcolor[HTML]{EAF0F6}42.35 & \cellcolor[HTML]{EAF0F6}42.22 \\
\bottomrule
\end{tabular}
}

%% file: analysis_data/sft/tab_sft_summary.tex
{\setlength{\tabcolsep}{12pt}\renewcommand{\arraystretch}{1.2}
\begin{tabular}{lcc}
\toprule
Model & Mean & HI \\
\midrule
Qwen3.8 27B (Base) & 63.19 & 42.78 \\
Qwen3.8 27B (SFT) & 68.30 & 46.86 \\
\midrule
\textit{Change} & \textbf{\boldmath $+$5.11} & \textbf{\boldmath $+$4.08} \\
\bottomrule
\end{tabular}}

%% file: analysis_data/sft/tab_sft_components.tex
{\setlength{\tabcolsep}{6pt}\renewcommand{\arraystretch}{1.08}
\begin{tabular}{@{}lrr!{\hspace{4pt}}rl@{}}
\toprule
Setting & Base & SFT & Change & \makebox[2.20cm]{\scriptsize\color{black!60}0\hfill $+$5\hfill $+$10} \\
\midrule \multicolumn{5}{@{}l}{\textit{Output protocol}} \\
\hspace{0.8em}Native tool calls & 63.98 & 73.48 & $+$9.50 & \makebox[0pt][l]{\color[HTML]{E8EDF3}\rule{2.20cm}{1.4ex}}{\color[HTML]{5E9C8F}\rule{2.089cm}{1.4ex}} \\
\hspace{0.8em}Python blocks & 64.76 & 66.87 & $+$2.11 & \makebox[0pt][l]{\color[HTML]{E8EDF3}\rule{2.20cm}{1.4ex}}{\color[HTML]{5E9C8F}\rule{0.464cm}{1.4ex}} \\
\hspace{0.8em}JSON blocks & 60.83 & 64.54 & $+$3.71 & \makebox[0pt][l]{\color[HTML]{E8EDF3}\rule{2.20cm}{1.4ex}}{\color[HTML]{5E9C8F}\rule{0.817cm}{1.4ex}} \\
\addlinespace[2pt] \multicolumn{5}{@{}l}{\textit{Action interface}} \\
\hspace{0.8em}Unified-schema API & 75.13 & 80.29 & $+$5.16 & \makebox[0pt][l]{\color[HTML]{E8EDF3}\rule{2.20cm}{1.4ex}}{\color[HTML]{5E9C8F}\rule{1.134cm}{1.4ex}} \\
\hspace{0.8em}Code-as-action & 64.58 & 67.60 & $+$3.02 & \makebox[0pt][l]{\color[HTML]{E8EDF3}\rule{2.20cm}{1.4ex}}{\color[HTML]{5E9C8F}\rule{0.665cm}{1.4ex}} \\
\hspace{0.8em}Primitive API & 49.83 & 57.02 & $+$7.20 & \makebox[0pt][l]{\color[HTML]{E8EDF3}\rule{2.20cm}{1.4ex}}{\color[HTML]{5E9C8F}\rule{1.583cm}{1.4ex}} \\
\addlinespace[2pt] \multicolumn{5}{@{}l}{\textit{Image budget}} \\
\hspace{0.8em}1 screenshot & 56.53 & 58.05 & $+$1.51 & \makebox[0pt][l]{\color[HTML]{E8EDF3}\rule{2.20cm}{1.4ex}}{\color[HTML]{5E9C8F}\rule{0.333cm}{1.4ex}} \\
\hspace{0.8em}5 screenshots & 68.52 & 75.31 & $+$6.79 & \makebox[0pt][l]{\color[HTML]{E8EDF3}\rule{2.20cm}{1.4ex}}{\color[HTML]{5E9C8F}\rule{1.494cm}{1.4ex}} \\
\hspace{0.8em}10 screenshots & 67.94 & 76.70 & $+$8.75 & \makebox[0pt][l]{\color[HTML]{E8EDF3}\rule{2.20cm}{1.4ex}}{\color[HTML]{5E9C8F}\rule{1.926cm}{1.4ex}} \\
\addlinespace[2pt] \multicolumn{5}{@{}l}{\textit{Text history}} \\
\hspace{0.8em}Full & 65.71 & 71.89 & $+$6.18 & \makebox[0pt][l]{\color[HTML]{E8EDF3}\rule{2.20cm}{1.4ex}}{\color[HTML]{5E9C8F}\rule{1.361cm}{1.4ex}} \\
\hspace{0.8em}Recent-$K$ & 64.66 & 70.22 & $+$5.56 & \makebox[0pt][l]{\color[HTML]{E8EDF3}\rule{2.20cm}{1.4ex}}{\color[HTML]{5E9C8F}\rule{1.224cm}{1.4ex}} \\
\hspace{0.8em}Operation briefs & 51.29 & 51.76 & $+$0.46 & \makebox[0pt][l]{\color[HTML]{E8EDF3}\rule{2.20cm}{1.4ex}}{\color[HTML]{5E9C8F}\rule{0.102cm}{1.4ex}} \\
\addlinespace[2pt] \multicolumn{5}{@{}l}{\textit{Input resolution}} \\
\hspace{0.8em}$1920\!\times\!1080$ & 54.39 & 59.28 & $+$4.89 & \makebox[0pt][l]{\color[HTML]{E8EDF3}\rule{2.20cm}{1.4ex}}{\color[HTML]{5E9C8F}\rule{1.077cm}{1.4ex}} \\
\hspace{0.8em}$1280\!\times\!720$ & 71.87 & 77.28 & $+$5.42 & \makebox[0pt][l]{\color[HTML]{E8EDF3}\rule{2.20cm}{1.4ex}}{\color[HTML]{5E9C8F}\rule{1.192cm}{1.4ex}} \\
\bottomrule
\end{tabular}}

%% file: tables/tab_hi_weight_check.tex
{\small
\setlength{\tabcolsep}{9pt}
\renewcommand{\arraystretch}{1.12}
\begin{tabular}{@{}lcc!{\hspace{4pt}}c}
\toprule
Model & Uniform configurations & Equal history weights & HI change \\
\midrule
\multicolumn{4}{@{}l}{\textit{Open-weight models}} \\
\hspace{0.6em}\raisebox{-0.22em}{\includegraphics[height=0.95em]{figure/component_scores/model_icons/glm.png}}\hspace{0.35em}GLM 5.3 Flash & 46.60 & 43.24 & \cellcolor[HTML]{D6DBE1}$-$3.36 \\
\hspace{0.6em}\raisebox{-0.22em}{\includegraphics[height=0.95em]{figure/component_scores/model_icons/deepseek.png}}\hspace{0.35em}DeepSeek v4.1 Flash & 43.73 & 39.61 & \cellcolor[HTML]{CBD2D8}$-$4.12 \\
\hspace{0.6em}\raisebox{-0.22em}{\includegraphics[height=0.95em]{figure/component_scores/model_icons/kimi.png}}\hspace{0.35em}Kimi K3 & 38.17 & 34.98 & \cellcolor[HTML]{D9DEE2}$-$3.19 \\
\hspace{0.6em}\raisebox{-0.22em}{\includegraphics[height=0.95em]{figure/component_scores/model_icons/kimi.png}}\hspace{0.35em}Kimi K2.6 & 12.31 & 10.98 & \cellcolor[HTML]{F3F4F6}$-$1.34 \\
\hspace{0.6em}\raisebox{-0.22em}{\includegraphics[height=0.95em]{figure/component_scores/model_icons/qwen.png}}\hspace{0.35em}Qwen3.8 Flash Next & 52.39 & 48.11 & \cellcolor[HTML]{C9D0D7}$-$4.28 \\
\hspace{0.6em}\raisebox{-0.22em}{\includegraphics[height=0.95em]{figure/component_scores/model_icons/qwen.png}}\hspace{0.35em}Qwen3.5 397B & 14.48 & 13.55 & \cellcolor[HTML]{F9F9FA}$-$0.93 \\
\midrule
\multicolumn{4}{@{}l}{\textit{Closed-source models}} \\
\hspace{0.6em}\raisebox{-0.22em}{\includegraphics[height=0.95em]{figure/component_scores/model_icons/openai.png}}\hspace{0.35em}GPT 6 Astra & 73.62 & 69.06 & \cellcolor[HTML]{C5CDD4}$-$4.56 \\
\hspace{0.6em}\raisebox{-0.22em}{\includegraphics[height=0.95em]{figure/component_scores/model_icons/claude.png}}\hspace{0.35em}Claude Fable 5 & 57.28 & 54.80 & \cellcolor[HTML]{E3E6EA}$-$2.47 \\
\bottomrule
\end{tabular}
}

%% file: appendix_action_interfaces.tex
\section{Action Interfaces and Output Protocols}
\label{app:action-contracts}

\definecolor{harnessink}{RGB}{38,44,54}
\definecolor{promptaccent}{RGB}{62,98,140}
\definecolor{promptbg}{RGB}{243,246,250}
\definecolor{callaccent}{RGB}{52,128,118}
\definecolor{callbg}{RGB}{244,248,247}
\definecolor{callstring}{RGB}{160,74,58}
\definecolor{callfence}{RGB}{150,156,164}
\definecolor{toolink}{RGB}{40,72,112}
\tcbset{harnessframe/.style={enhanced,sharp corners,frame hidden,
  borderline west={1.8pt}{0pt}{#1},boxrule=0pt,
  left=8pt,right=8pt,top=1pt,bottom=4pt,toptitle=3.5pt,bottomtitle=0pt,
  titlerule=0pt,fonttitle=\sffamily\scriptsize\bfseries,coltitle=#1,
  before skip=6pt,after skip=2pt}}
\newtcolorbox{harnessprompt}[1][System prompt]{harnessframe=promptaccent,
  colback=promptbg,colbacktitle=promptbg,title={\MakeUppercase{#1}},
  before upper={\setlength{\parskip}{2pt}\setlength{\parindent}{0pt}},
  fontupper=\fontencoding{T1}\selectfont\small\sffamily,colupper=harnessink}
\newcommand{\promptnote}[1]{\par\noindent{\small\color{black!70}\textit{#1}}\par\smallskip}
\lstdefinestyle{harnesscall}{morestring=[b]',morestring=[b]",
  basicstyle=\fontencoding{T1}\selectfont\small\ttfamily\color{harnessink},
  columns=fullflexible,keepspaces=true,showstringspaces=false,upquote=true,
  breaklines=true,breakatwhitespace=false,
  stringstyle=\color{callstring},
  morecomment=[l]{```},commentstyle=\color{callfence},
  emph={Action,Native,Python,JSON},emphstyle=\color{promptaccent}\bfseries,
  aboveskip=0pt,belowskip=0pt}
\newtcblisting{harnesscall}[1][Example response]{harnessframe=callaccent,
  colback=callbg,colbacktitle=callbg,title={\MakeUppercase{#1}},
  listing only,listing options={style=harnesscall}}
\newcommand{\contracttablesetup}{\renewcommand{\arraystretch}{1.2}}
\newcommand{\contracttop}{\arrayrulecolor{promptaccent}\toprule\arrayrulecolor{black!40}}
\newcommand{\contractbottom}{\arrayrulecolor{promptaccent}\bottomrule\arrayrulecolor{black}}
\newcommand{\contractsep}{\arrayrulecolor{black!14}\specialrule{0.3pt}{0pt}{0pt}}
\newcommand{\headcell}[1]{\textcolor{promptaccent}{\textbf{#1}}}
\newcolumntype{T}[1]{>{\raggedright\arraybackslash\leavevmode\color{toolink}}p{#1}}

\subsection{Overview}
\label{app:interface-overview}
Each configuration combines an \emph{action interface} with an \emph{output protocol}.
The action interface specifies the tools available to the model and their arguments; the output protocol specifies how the model writes these tool calls in its response.
Each of the three interfaces supports three output protocols (native tool calls, Python code blocks, and JSON code blocks), giving nine combinations.
The interface names describe the form of the action space and do not restrict which models may use them.
Tab.~\ref{tab:action-contract-overview} summarizes the main differences among the three interfaces.

\begin{table}[!b]
\centering\small
\caption{Overview of the three action interfaces. All three accept an ordered sequence of calls in a single decision. Pixel coordinates refer to the displayed screenshot, whose resolution may differ from that of the desktop.}
\label{tab:action-contract-overview}
\setlength{\tabcolsep}{5pt}\contracttablesetup
\begin{tabular}{@{}>{\raggedright\arraybackslash}p{0.19\linewidth}>{\raggedright\arraybackslash}p{0.25\linewidth}>{\raggedright\arraybackslash}p{0.24\linewidth}>{\raggedright\arraybackslash}p{0.25\linewidth}@{}}
\contracttop
 & \headcell{Primitive API} & \headcell{Unified-schema API} & \headcell{Code-as-action} \\
\midrule
Tools & Separate typed tools & One \texttt{computer\_use} tool & Python program plus auxiliary tools \\
\contractsep
Position format & Separate integer \texttt{x}, \texttt{y} & Integer \texttt{coordinate=[x,y]}, each in $[0,999]$ & Image-pixel positions or relative displacements \\
\contractsep
Composition & Ordered tool calls & Ordered action-discriminated calls & Ordered tool calls; variables and loops inside \texttt{code} \\
\contractsep
Completion & \texttt{finish}: \texttt{done/failed} & \texttt{terminate}: \texttt{success/failure}; or \texttt{answer} & \texttt{finish\_task} or \texttt{computer(terminate)} \\
\contractbottom
\end{tabular}
\end{table}

\paragraph{System prompt.}
The system prompt consists of four parts: shared instructions, the description and argument schema of the selected interface, the instructions for the selected output protocol, and a response example.
For native tool calls, the same schemas are also passed to the model API as tool definitions.
The task instruction, screenshots, and interaction history are provided outside the system prompt, which does not describe how history is retained.
The box below shows selected paragraphs of the shared instructions; the code-as-action prompt in Sec.~\ref{app:code-contract} is shown for a $1280\!\times\!720$ screenshot.

\begin{harnessprompt}[System prompt \textperiodcentered{} shared]
\input{analysis_data/action_interfaces/prompt_common.tex}
\end{harnessprompt}

\subsection{Primitive API}
\label{app:primitive-contract}
The primitive API provides a separate tool for each operation.
For a screenshot of width $W$ and height $H$, \texttt{click} and \texttt{move} take two integer arguments with $0\leq x<W$ and $0\leq y<H$.
These coordinates are mapped to the $1920\!\times\!1080$ desktop as
$(\lfloor1920x/W\rfloor,\lfloor1080y/H\rfloor)$.
The model therefore works directly in screenshot pixels and does not normalize or rescale coordinates.

\begin{table}[htbp]
\centering\small
\caption{Tools of the primitive API. Square brackets denote optional arguments, shown with their default values.}
\label{tab:primitive-tool-contract}
\contracttablesetup
\begin{tabular}{@{}T{0.31\linewidth}>{\raggedright\arraybackslash}p{0.64\linewidth}@{}}
\contracttop
\headcell{Tool(s)} & \headcell{Arguments and semantics} \\
\midrule
\texttt{click}; \texttt{move} & \texttt{x, y}; click also accepts [\texttt{button=left}], [\texttt{count=1}]. Count is 1--3. \\
\contractsep
\texttt{drag} & \texttt{start=\{x,y\}, end=\{x,y\}}; [\texttt{button=left}], [\texttt{seconds=0.5}]. Duration: 0.05--5 s. \\
\contractsep
\texttt{type\_text} & \texttt{text}; ASCII is typed, other text is pasted through the clipboard. \\
\contractsep
\texttt{press}; \texttt{hotkey} & \texttt{key}; or a \texttt{keys} list with 1--8 declared key names. \\
\contractsep
\texttt{key\_down}; \texttt{key\_up} & \texttt{key}; held state persists until released. \\
\contractsep
\texttt{mouse\_down}; \texttt{mouse\_up} & \texttt{button} in \texttt{left/right/middle}; held state persists. \\
\contractsep
\texttt{scroll}; \texttt{wait} & Scroll: \texttt{dy}, [\texttt{dx=0}], each in $[-100,100]$ wheel notches. Wait: \texttt{seconds} in $[0,60]$. \\
\contractsep
\texttt{finish} & \texttt{status} in \texttt{done/failed}; terminates for task grading. \\
\contractbottom
\end{tabular}
\end{table}

\begin{harnessprompt}[Interface prompt \textperiodcentered{} Primitive API]
\input{analysis_data/action_interfaces/prompt_primitive.tex}
\end{harnessprompt}

\paragraph{Example.}
The following response in the Python-block protocol first moves the cursor to a visible region and then scrolls.
The two lines are parsed into two tool calls; the code block itself is not executed.

\begin{harnesscall}
Action: Move to the document area and scroll down.
```python
move(x=240, y=160)
scroll(dy=-3)
```
\end{harnesscall}

\paragraph{Argument validation.}
Each coordinate must be an integer; values such as \texttt{x=[240]} or numeric strings are rejected.
Unknown arguments and undefined tools are also rejected; for example, a double click is written as \texttt{click(count=2)} rather than as a separate \texttt{doubleClick} tool.
All calls in a response are validated before any of them is executed.
A single decision may therefore contain several actions, even under this interface.

\subsection{Unified-schema API}
\label{app:unified-contract}
The unified-schema API provides a single tool, \texttt{computer\_use}, whose \texttt{action} argument selects the operation.
Each action has its own set of valid arguments, so an argument accepted by one action may be rejected by another.
Positions are given as a two-element integer array on a fixed $[0,999]^2$ grid and mapped to the desktop as
$(\lfloor1919x/999\rfloor,\lfloor1079y/999\rfloor)$.
This representation does not depend on the screenshot resolution.

\begin{table}[htbp]
\centering\small
\caption{Actions of the \texttt{computer\_use} tool in the unified-schema API. For click actions, the optional \texttt{text} argument specifies held modifier keys rather than text to be typed.}
\label{tab:unified-tool-contract}
\contracttablesetup
\begin{tabular}{@{}T{0.39\linewidth}>{\raggedright\arraybackslash}p{0.56\linewidth}@{}}
\contracttop
\headcell{Action value(s)} & \headcell{Arguments and semantics} \\
\midrule
\texttt{mouse\_move}; \texttt{left\_click\_drag} & \texttt{coordinate=[x,y]}; drag starts at the current cursor and lasts 0.5 s. \\
\contractsep
\texttt{left\_click}, \texttt{right\_click}, \texttt{middle\_click}, \texttt{double\_click}, \texttt{triple\_click} & \texttt{coordinate}; optional \texttt{text} holds \texttt{ctrl/shift/alt/win} variants joined by \texttt{+}. \\
\contractsep
\texttt{key}; \texttt{type} & \texttt{keys} list; or \texttt{text} to type/paste. \\
\contractsep
\texttt{scroll}; \texttt{hscroll} & \texttt{pixels} in $[-100,100]$; optional modifier \texttt{text}. Positive means up/right. No \texttt{coordinate} field. \\
\contractsep
\texttt{wait} & Optional \texttt{time} in $[0,60]$ seconds; default 1 s. \\
\contractsep
\texttt{terminate}; \texttt{answer} & \texttt{status=success/failure}; or \texttt{text} as a final answer. \\
\contractbottom
\end{tabular}
\end{table}

\begin{harnessprompt}[Interface prompt \textperiodcentered{} Unified-schema API]
\input{analysis_data/action_interfaces/prompt_unified.tex}
\end{harnessprompt}

\paragraph{Example.}
The same move-and-scroll operation requires two calls to \texttt{computer\_use} with different actions.
The coordinates below are on the normalized grid and do not correspond to the pixel coordinates in the primitive-API example.

\begin{harnesscall}
Action: Move to the document area and scroll down.
```python
computer_use(action='mouse_move', coordinate=[250, 250])
computer_use(action='scroll', pixels=-3)
```
\end{harnesscall}

\paragraph{Differences from the primitive API.}
The \texttt{pixels} argument of \texttt{scroll} counts wheel steps rather than screen pixels.
Because \texttt{scroll} takes no coordinate, scrolling a specific region requires a preceding \texttt{mouse\_move} call.

\subsection{Code-as-action Interface}
\label{app:code-contract}
Under code-as-action, the model writes a Python program using PyAutoGUI~\citep{pyautogui_docs} and passes it as the \texttt{code} argument of \texttt{execute\_pyautogui\_code}.
Unlike the Python-block protocol, in which code blocks are only parsed, this program is executed on the desktop and may contain variables, loops, and multiple GUI operations.
Before execution, the program is checked for syntax errors and for the argument signatures of scroll calls; other functions are not restricted by an allowlist.

\begin{table}[htbp]
\centering\small
\caption{Tools of the code-as-action interface. Only one screen (\texttt{:0}) is available; \texttt{screen\_id} is optional where supported.}
\label{tab:code-tool-contract}
\contracttablesetup
\begin{tabular}{@{}T{0.35\linewidth}>{\raggedright\arraybackslash}p{0.60\linewidth}@{}}
\contracttop
\headcell{Tool} & \headcell{Arguments and semantics} \\
\midrule
\texttt{execute\_pyautogui\_code} & Nonblank \texttt{code} string, at most 20,000 characters; [\texttt{screen\_id=:0}]. \texttt{pyautogui} is imported. \\
\contractsep
\texttt{take\_screenshot} & [\texttt{screen\_id=:0}]; observes without an input action. \\
\contractsep
\texttt{wait\_for\_state} & [\texttt{seconds=20}] in 1--60; [\texttt{reason}]. A timed wait, not an automatically tested state predicate. \\
\contractsep
\texttt{computer} & \texttt{action=wait}: waits 1 s; or \texttt{action=terminate}, required \texttt{status=success/failure}, optional \texttt{answer}. \\
\contractsep
\texttt{finish\_task} & Required \texttt{status=success/failure}; optional \texttt{answer}. \\
\contractbottom
\end{tabular}
\end{table}

\begin{harnessprompt}[Interface prompt \textperiodcentered{} Code-as-action]
\input{analysis_data/action_interfaces/prompt_code.tex}
\end{harnessprompt}

\paragraph{Example.}
The outer call contains only keyword arguments with literal values, and the loop is written inside the \texttt{code} string.
Both tool calls belong to the same decision.

\begin{harnesscall}
Action: Focus the list, move down twice, and wait.
```python
execute_pyautogui_code(code='pyautogui.click(x=240, y=160)\nfor i in range(2):\n    pyautogui.press("down")')
wait_for_state(seconds=2)
```
\end{harnesscall}

\noindent Pointer positions and relative movements in the program are given in screenshot pixels and are rescaled to desktop coordinates for the supported PyAutoGUI functions; other imported functions are not rescaled.

\subsection{Output Protocols}
\label{app:output-prompts}
The output protocol determines how tool calls are written in the response; it does not change the available tools.
The instructions for native tool calls are shown in full; for the Python- and JSON-block protocols, selected paragraphs are shown.

\medskip\noindent\textbf{Native tool calls.}\par
\begin{harnessprompt}[Output protocol \textperiodcentered{} Native tool calls]
\input{analysis_data/action_interfaces/prompt_native.tex}
\end{harnessprompt}

\noindent\textbf{Python code blocks.}\par
\begin{harnessprompt}[Output protocol \textperiodcentered{} Python block]
\input{analysis_data/action_interfaces/prompt_python.tex}
\end{harnessprompt}

\noindent\textbf{JSON code blocks.}\par
\begin{harnessprompt}[Output protocol \textperiodcentered{} JSON block]
\input{analysis_data/action_interfaces/prompt_json.tex}
\end{harnessprompt}

\paragraph{One call in three protocols.}
The example below writes the same unified-schema API call in each protocol.
Under native tool calls, the call is returned in the tool-call field of the API response; in the Python- and JSON-block protocols, it follows the \texttt{Action:} line in the message text.

\begin{harnesscall}[Same call in three protocols]
Native: function.name = "computer_use"
        function.arguments = '{"action":"left_click","coordinate":[250,250]}'
Python:
```python
computer_use(action='left_click', coordinate=[250, 250])
```
JSON:
```json
{"action_type":"computer_use",
 "parameters":{"action":"left_click","coordinate":[250,250]}}
```
\end{harnesscall}

\paragraph{Parsing and validation.}
Each response is parsed according to the selected protocol, and the complete call sequence is validated before execution.
In the standard setting, the model has up to three attempts in total to produce a valid response, using the unchanged request without error messages; the feedback experiments instead return the validation error to the model.
All examples in this section pass the parser and validator at both screenshot resolutions; this check confirms that they are well formed, but does not execute them.

%% file: analysis_data/action_interfaces/prompt_common.tex
You operate a desktop to complete the user's task. Use the declared tools.

Choose one tool call or an ordered sequence of tool calls per decision. Calls execute sequentially in the returned order, never in parallel. You receive the next observation after the sequence. A code tool may contain multiple consecutive GUI operations. A completion call must be last.

Use visible GUI interaction for the task. Do not inspect or modify evaluator files. Do not read hidden task metadata.

Every call in a sequence must satisfy the selected format and tool schema before any call executes. Do not assume a requested operation succeeded; inspect the resulting screen.

Each Action: description states what you do and to which visible element, for example \textquotedbl{}Click the Save button\textquotedbl{} or \textquotedbl{}Type the sheet name into the Name field\textquotedbl{}, not coordinates.

%% file: analysis_data/action_interfaces/prompt_primitive.tex
Separate click, move, drag, type\_text, press, hotkey, key\_down/up, mouse\_down/up, scroll, wait and finish tools. Coordinates are absolute pixel positions in the displayed image, with origin (0,0) at the top-left, x increasing rightward and y downward. They must be inside that image; the executor maps them to the desktop. Do not scale them yourself or use normalized coordinates. click defaults left/1; drag defaults left/0.5 seconds. type\_text types ASCII and pastes other text. click takes separate integer x and y fields, for example \{\textquotedbl{}x\textquotedbl{}:240,\textquotedbl{}y\textquotedbl{}:160\}. Double-click uses click with count=2; triple-click uses count=3; right-click uses button=\textquotedbl{}right\textquotedbl{}. There are no doubleClick, triple\_click or rightClick tools. scroll uses wheel notches: dx positive right, dy positive up, dx defaults to 0. It scrolls at the current mouse position; to scroll a specific area, first move there. mouse\_down and mouse\_up require button. Held keys and buttons persist until released. Use exact declared key names; the system key is win, not super; command/option are not supported on Ubuntu. finish requires done/failed.

%% file: analysis_data/action_interfaces/prompt_unified.tex
Use computer\_use(action=...). Coordinates are normalized positions, not cursor-relative displacements: integer [x,y] in [0,999] on each axis, with 0 at the top/left and 999 at the bottom/right. x increases rightward and y downward. The grid stays the same for either displayed-image size. key uses a key list. type types ASCII and pastes other text. Clicks require a coordinate; optional text holds modifiers separated by + (e.g. ctrl+shift). left\_click\_drag starts at the current cursor and drags to coordinate in 0.5 seconds. triple\_click clicks three times. scroll moves vertically, hscroll horizontally; pixels means signed wheel notches, positive up/right. scroll/hscroll act at the current cursor and accept only action, pixels and optional text; coordinate is not accepted. To scroll at a location, first call mouse\_move with coordinate, then scroll with pixels. wait defaults to 1 second. terminate requires success/failure; answer completes with text.
The action-specific anyOf branches below define the required and allowed fields. Modifier text holds only ctrl, shift, alt or win (or their left/right variants) separated by +; to press + itself use action=key. The system key is win, not super; macOS command/option names are not supported on this Ubuntu desktop.

%% file: analysis_data/action_interfaces/prompt_code.tex
execute\_pyautogui\_code takes Python code with pyautogui imported. The image coordinate canvas is 1280x720 pixels. click/moveTo/dragTo use absolute pixel positions in the displayed image, with origin (0,0) at the top-left, x increasing rightward and y downward. moveRel/dragRel and their move/drag aliases use displacements from the current cursor in displayed-image pixels, positive right/down. An omitted relative axis means zero movement on that axis. The executor maps positions and displacements to desktop pixels; do not scale them yourself. pyautogui.size(), pyautogui.resolution() and pyautogui.position() use the same image coordinate system. Screen ID defaults to :0, the only screen. Consecutive GUI operations may be batched in one code call, including variables and loops. Code must be nonblank and syntactically valid. Static scroll signatures are checked: scroll/hscroll/vscroll(clicks, x=None, y=None, logScreenshot=None, \_pause=True). This is not a strict function whitelist; other Python/runtime errors may occur during execution. Coordinate adaptation covers pointer and relative movement, pixel queries, position/size, screenshots and locateOnScreen/locateCenterOnScreen regions and results. Regions and boxes are (left,top,width,height) in image pixels. Pointer targets can also be a point, a box to click at its center, or a desktop-template filename. Saved screenshots use the same dimensions as returned images. Screen matching still uses desktop-resolution template images; other helpers and direct imports retain their library coordinate systems. Use code for GUI interaction, not direct filesystem, shell, network, or evaluator access. take\_screenshot observes without acting. wait\_for\_state defaults to 20 seconds. computer(action=wait) waits 1 second and accepts no status/answer; computer(action=terminate) requires status and permits answer. finish\_task requires success/failure and permits answer. No run\_command or ask\_user tool is provided.

%% file: analysis_data/action_interfaces/prompt_native.tex
Write one Action: line with a short description of this decision in the message content, and return one or more API function tool calls using the registered tools. The tool\_calls list defines execution order; the executor runs calls sequentially. Do not encode tool invocations as JSON or XML in message content.

%% file: analysis_data/action_interfaces/prompt_python.tex
Start with one Action: line with a short description of this decision, then write the tool calls inside a \textasciigrave{}\textasciigrave{}\textasciigrave{}python code block. Each line is one call to a declared tool, written as a Python function call with keyword arguments only.

Argument values must be literals: strings, numbers, True/False/None, lists (tuples count as lists) or dicts; values cannot contain \textasciigrave{}\textasciigrave{}\textasciigrave{}; no variables, expressions, imports or other statements. These rules apply to the call lines; the content of a string argument, such as Python code passed to a code tool, is not restricted by them.

The harness parses the blocks without executing them and validates every call before execution.

%% file: analysis_data/action_interfaces/prompt_json.tex
Start with one Action: line with a short description of this decision, then write the tool calls inside a \textasciigrave{}\textasciigrave{}\textasciigrave{}json code block. Each call is a JSON object \{\textquotedbl{}action\_type\textquotedbl{}: \textquotedbl{}\textless{}declared tool name\textgreater{}\textquotedbl{}, \textquotedbl{}parameters\textquotedbl{}: \{\textless{}tool arguments\textgreater{}\}\}; a block may hold one object or a nonempty array of objects.

Several json blocks form an ordered sequence. Explanations may appear outside the blocks; every fenced block in the response must be a json block. Every call is validated before execution. Incomplete or invalid blocks are not skipped or repaired.